\documentclass[lettersize,journal]{IEEEtran}
\usepackage{amsmath,amsfonts}
\usepackage{algorithmic}
\usepackage{algorithm}
\usepackage{array}
\usepackage[caption=false,font=normalsize,labelfont=sf,textfont=sf]{subfig}
\usepackage{textcomp}
\usepackage{stfloats}
\usepackage{url}
\usepackage{verbatim}
\usepackage{graphicx}
\usepackage{cite}
\usepackage{threeparttable}
\usepackage{color}
\usepackage{wrapfig}
\usepackage{booktabs}
\usepackage{multirow}
\usepackage{capt-of}
\usepackage{enumerate}
\usepackage{makecell} %
\usepackage{color}%
\usepackage{colortbl} %
\usepackage[dvipsnames, svgnames, x11names]{xcolor}
\usepackage{mathrsfs} %
\usepackage{amssymb} %
\usepackage{xcolor}
\usepackage{pifont}%
\definecolor{eva01}{HTML}{A877C8}

\usepackage{xspace}
\usepackage{mathabx}

\usepackage[colorlinks=true]{hyperref}
\makeatletter
\DeclareRobustCommand\onedot{\futurelet\@let@token\@onedot}
\def\@onedot{\ifx\@let@token.\else.\null\fi\xspace}
 
\def\ie{\emph{i.e}\onedot}

\makeatother

\newcommand{\name}{WeakTr}

\newcommand{\revision}[1]{#1}

\begin{document}

\title{\name: Exploring Plain Vision Transformer for Weakly-supervised Semantic Segmentation}

\author{
    Lianghui Zhu, Yingyue Li, Jiemin Fang, Yan Liu, Xin Hao, Wenyu Liu,~\IEEEmembership{Senior Member,~IEEE}, and Xinggang Wang$^\dagger$,~\IEEEmembership{Senior Member,~IEEE}
    \IEEEcompsocitemizethanks{
        \IEEEcompsocthanksitem L. Zhu, Y. Li, J. Fang, W. Liu and X. Wang are with the School of Electronic Information and Communications, Huazhong University of Science and Technology, Wuhan 430074, P.R. China.
        \IEEEcompsocthanksitem  Y. Liu and X. Hao are with Alipay Tian Qian Security Lab.
    }
    \thanks{$^\dagger$ Corresponding to X. Wang (\url{xgwang@hust.edu.cn}).}
}

\markboth{Journal of \LaTeX\ Class Files,~Vol.~14, No.~8, August~2021}%
{Shell \MakeLowercase{\textit{et al.}}: A Sample Article Using IEEEtran.cls for IEEE Journals}

\maketitle

\begin{abstract}
  Transformer has been very successful in various computer vision tasks and understanding the working mechanism of transformer is important. As touchstones, weakly-supervised semantic segmentation (WSSS) and class activation map (CAM) are useful tasks for analyzing vision transformers (ViT). Based on the plain ViT pre-trained with ImageNet classification,
  we find that multi-layer, multi-head self-attention maps can provide rich and diverse information for weakly-supervised semantic segmentation and CAM generation, e.g., different attention heads of ViT focus on different image areas and object categories. Thus we propose a novel method to end-to-end estimate the importance of attention heads, where the self-attention maps are adaptively fused for high-quality CAM results that tend to have more complete objects. Besides, we propose a ViT-based gradient clipping decoder for online retraining with the CAM results efficiently and effectively. Furthermore, the gradient clipping decoder can make good use of the knowledge in large-scale pre-trained ViT and has a scalable ability. The proposed plain \underline{\textbf{Tr}}ansformer-based \underline{\textbf{Weak}}ly-supervised learning method (\name{}) obtains the superior WSSS performance on standard benchmarks, 
  \ie, 
  78.5\% mIoU on the $val$ set of PASCAL VOC 2012 and 51.1\% mIoU on the $val$ set of COCO 2014. Source code and checkpoints are available at \url{https://github.com/hustvl/WeakTr}.
\end{abstract}


\section{Introduction}\label{sec:introduction}
\IEEEPARstart{W}{eakly-supervised} semantic segmentation (WSSS) aims to alleviate the reliance on pixel-level semantic annotations by utilizing weak annotations~\cite{shen2023re1,si2024re3}. Among them, only using image-level class labels is the most challenging. Due to the lack of positional annotations, image-level WSSS methods usually require coarse position annotations generated by the class activation map (CAM) \cite{zhou2016cam}.
Given the pervasive application of CAM across various domains within deep learning, improving CAM is imperative. Its enhancement not only substantially bolsters WSSS but also holds profound significance in the arenas of model interpretability~\cite{selvaraju2017gradcam} and network regularization~\cite{lin2021camregularization}, among others.
To improve CAM for higher-quality pseudo mask generation, most previous WSSS frameworks \cite{chang2020sccamweakly, zhang2020causal, xie2022clims, xu2022mctformer} introduced the CAM refinement phase \cite{ahn2018psa, ahn2019irn}. These pseudo masks are further used for supervising the segmentation networks \cite{liang2015dplabv1, chen2017dplabv2} in a retraining phase.

\begin{figure}[t]
  \subfloat[\footnotesize{Averaged attention values \\ on different patches.}]{
      \begin{minipage}{0.47 \linewidth}
          \includegraphics[width=\linewidth]{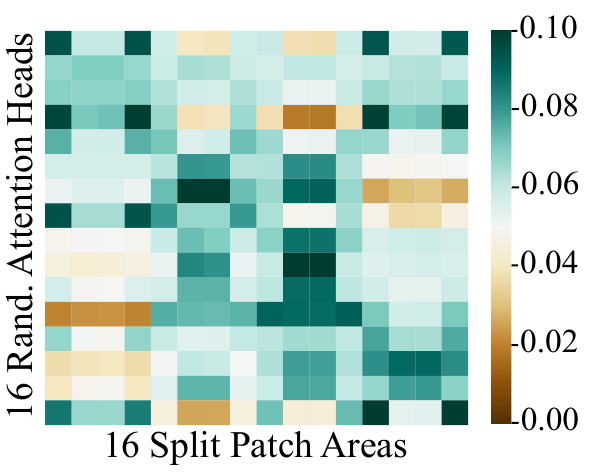}
  
          \label{fig:attn_heads_to_areas}
      \end{minipage}
  }
  \subfloat[\footnotesize{PCC between averaged attention and class labels.}]{
      \begin{minipage}{0.47 \linewidth}
          \includegraphics[width=\linewidth]{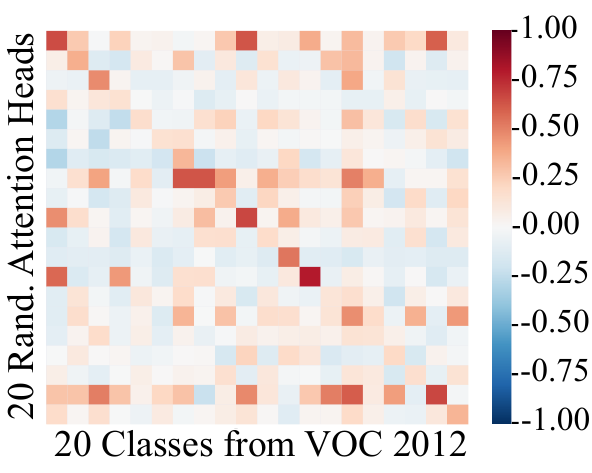}
  
          \label{fig:attn_heads_to_classes}
      \end{minipage}
  }
  \caption{The statistical relationship between the attention values of different ViT heads and the patch position \& the image category based on the whole VOC 2012 $train$ set. In (a), we average along the class dimension and observe their attention values at spatial positions.
  In (b), we average along the spatial dimension and compute the PCC~\cite{cohen2009pcc} between the averaged values and the classification ground truth.}
  \label{fig:statistical_results}
\vspace{-0.5cm}
\end{figure}

With the success of vision transformers (ViT) \cite{dosovitskiy2020vit}, some methods~\cite{xu2022mctformer, gao2021tscam} propose to obtain CAM seeds with the assistance of transformer features and corresponding self-attention maps. 
The CAM generation phase usually contains 2 stages. In the first stage, they generate coarse CAM through feature tokens. Self-attention maps, representing relations between feature tokens, are then adopted to enhance the coarse CAM in the post-processing stage. 
These methods directly average the attention maps across different heads and sum them by layer. However, as shown in Fig.~\ref{fig:statistical_results}, most different attention heads focus on different positions and object categories, which may contain information unrelated to the target object. Direct averaging and summing may lead to misleading information in the post-processing stage. 
We propose an adaptive attention fusion module (AAF) to measure the importance of different attention heads for CAM, which is used for assigning weights to attention heads. To ensure that AAF can estimate accurate weights, we propose an end-to-end training strategy for CAM generation. During the training process, we apply the weights estimated by the AAF module to the attention heads. The coarse CAM is thus optimized to be finer with the weighted self-attention maps. 
The way AAF assigns weight for different attention heads is similar to the ViT mechanism that assigns weight for different feature tokens. As a weight supplement to attention heads, the AAF brings improvements for both CAM quality and optimization.

The success of ViT relies on the emergence of a large number of pre-training methods, which include techniques such as self-supervised methods~\cite{oquab2023dinov2}, strong data augmentations~\cite{touvron2021deit}, the utilization of large-scale pre-training data~\cite{steiner2021improvevit}, and text supervision~\cite{fang2023eva, fang2023eva01, sun2023evaclip}. It is also important for WSSS to harness the powerful representations from pre-trained ViTs. Besides, the pre-trained ViTs involving learning from large-scale pre-training data can capture rich semantic representations that are more robust than hand-crafted WSSS priors (e.g., Random Walk~\cite{ahn2019irn}).
So, we explore a large-scale pre-trained ViT-based method to perform online retraining on CAM seeds without the CAM refinement phase.
Intuitively, regions with wrong labels in CAM will affect the training of segmentation networks. 
Motivated by methods \cite{han2018coteacher, arazo2019losscorr, ren2018reweight} for noisy label problems in classification networks, we propose a gradient clipping decoder for identifying confident regions. More specifically, image areas with a larger gradient are filtered out by a gradient clipping decoder. In this way, the segmentation network tends to be updated with smaller gradients for confident CAM regions. Our online retraining method enables the segmentation network to efficiently learn CAM regions that are biased towards correct labeling. 
Furthermore, the proposed gradient clipping decoder can effectively harness the power of the large-scale pre-trained ViTs and exhibit scalable performance as the pre-trained data increases.
Note that our online retraining network can not only be a high-performance segmentation network itself, but also be used as a labeling tool for producing high-quality pseudo labels for other segmentation networks.

The main contributions of this paper can be summarized as follows:

\begin{itemize}
    \item We exploit the inherent properties of multi-layer, multi-head self-attention maps in plain ViT and devise an effective adaptive attention fusion strategy for generating high-quality class activation maps. This is the first work that sheds light on the importance of different attention heads for CAM and WSSS.

    \item We have explored a concise and efficient framework (WeakTr) based on plain pre-trained ViTs for WSSS. Our WeakTr framework enables end-to-end generation of high-quality CAM and efficient online retraining through a simple but effective gradient clipping decoder. The overall training speed of the WeakTr framework is about 2.6 times that of the baseline framework.
    
    \item Our \name{} fully explores the potential of plain ViT in the WSSS domain. Superior results are achieved on both challenging WSSS benchmarks, with 78.5\% mIoU on PASCAL VOC 2012~\cite{everingham2010voc12} and 51.1\% on COCO 2014~\cite{lin2014coco14} validation sets, respectively, significantly surpassing previous methods.
    
\end{itemize}

\section{Related Work}
\label{Sec: RelatedWork}
    \subsection{Transformer in WSSS}
    The vision transformer has recently advanced the field of computer vision. \revision{Plain} ViT \cite{dosovitskiy2020vit} transforms images into non-overlapping patch tokens, which are used as input along with a class token. The class token is then mapped to a class prediction using a fully connected layer. 
    \revision{Plain ViT refers to a vanilla, non-hierarchical Vision Transformer encoder.}
    This model architecture without convolutional induction bias is considered to be promising. The TS-CAM \cite{gao2021tscam} method uses the cross-attention map between the class token and patch tokens to obtain location cues in the weakly-supervised domain. The acquisition of the cross-attention map requires averaging the attention maps of different heads under the same layer and then summing over the different layers. After this, the cross-attention map is combined with the CAM obtained by processing the patch tokens using convolution. After this method, MCTformer \cite{xu2022mctformer} proposes multiple class tokens as input for learning the cross-attention maps of different classes. The CAM is additionally optimized by using the patch-attention maps in the post-processing stage. In addition, TransCAM \cite{li2022transcam} is based on the conformer \cite{peng2021conformer} backbone, which is a mixture of transformer blocks and convolution. It also uses patch-attention maps to refine CAM at the CAM generation stage. However, most attention heads of the transformer notice different positions and classes in the image, which may contain information unrelated to the target object.

    Unlike the above ViT-based methods, our WeakTr uses different weights to estimate the importance of the transformer's attention heads. With an end-to-end strategy to optimize the adaptive attention fusion module, we could further improve the accuracy of the final attention result.

    \subsection{Image-level Supervised Learning}

    In order to obtain cues with only image-level labels, many methods focus on how to optimize CAM. The SEC method \cite{kolesnikov2016sec} spreads the sparse CAM labels by seed expansion. \revision{DSRG \cite{huang2018dsrg} and Usage~\cite{peng2023re2} combine the seed region growth method to expand CAM cues}. A similar approach is DGCN \cite{feng2021dgcn}, which assigns labels to regions around seeds by using traditional graph-cutting algorithms. AffinityNet \cite{ahn2018psa} and IRNet \cite{ahn2019irn} propagate the labels using the random walk method. AuxSegNet \cite{xu2021auxsegnet} propagates labels by learning the affinity of the cross-task. There are also methods that use adversarial erasing \cite{hou2018selferasing,wei2017adverasing} to help CAM focus more on the undiscriminating regions. SEAM \cite{wang2020seam} explores the consistency of CAM under different affine transformations. In addition, there are methods that choose to introduce web data, such as Co-segmentation \cite{shen2017coseg} and STC \cite{wei2016stc}.

    However, using methods such as AffinityNet \cite{ahn2018psa} to refine the CAM and then using the refined pseudo mask to retrain the DeepLab \cite{liang2015dplabv1,chen2017dplabv2} network can be too complicated and time-consuming. Our proposed gradient clipping decoder enables the segmentation network to directly and efficiently learn from confident CAM areas without any CAM refinement.

\section{Method}
\label{sec: Method}
\name~includes two phases, end-to-end CAM generation, and online retraining. In this section, we first introduce the end-to-end CAM generation phase of the WeakTr framework, which comprises a plain ViT backbone and an adaptive attention fusion module for generating fine CAM end-to-end. We then discuss the online retraining phase of the WeakTr framework, which also employs the plain ViT backbone and a gradient clipping decoder that enables direct retraining without CAM refinement.

\subsection{CAM Generation Framework of \name}

The CAM generation consists of a plain ViT backbone, the adaptive attention fusion (AAF) module, CAM generation heads, and end-to-end training. At first, we introduce the architecture of the ViT backbone and the key mechanism, multi-head self-attention. Next, we perform adaptive attention fusion (AAF) to produce proper weights for attention heads. Then, we optimize the coarse CAM with weighted self-attention maps. Finally, the proposed end-to-end training updates the parameters of WeakTr, and especially the AAF module, with image-level supervision.

\subsubsection{Plain ViT Backbone}

\begin{figure*}[t]
    \centering
    \includegraphics[width=1.0\linewidth]{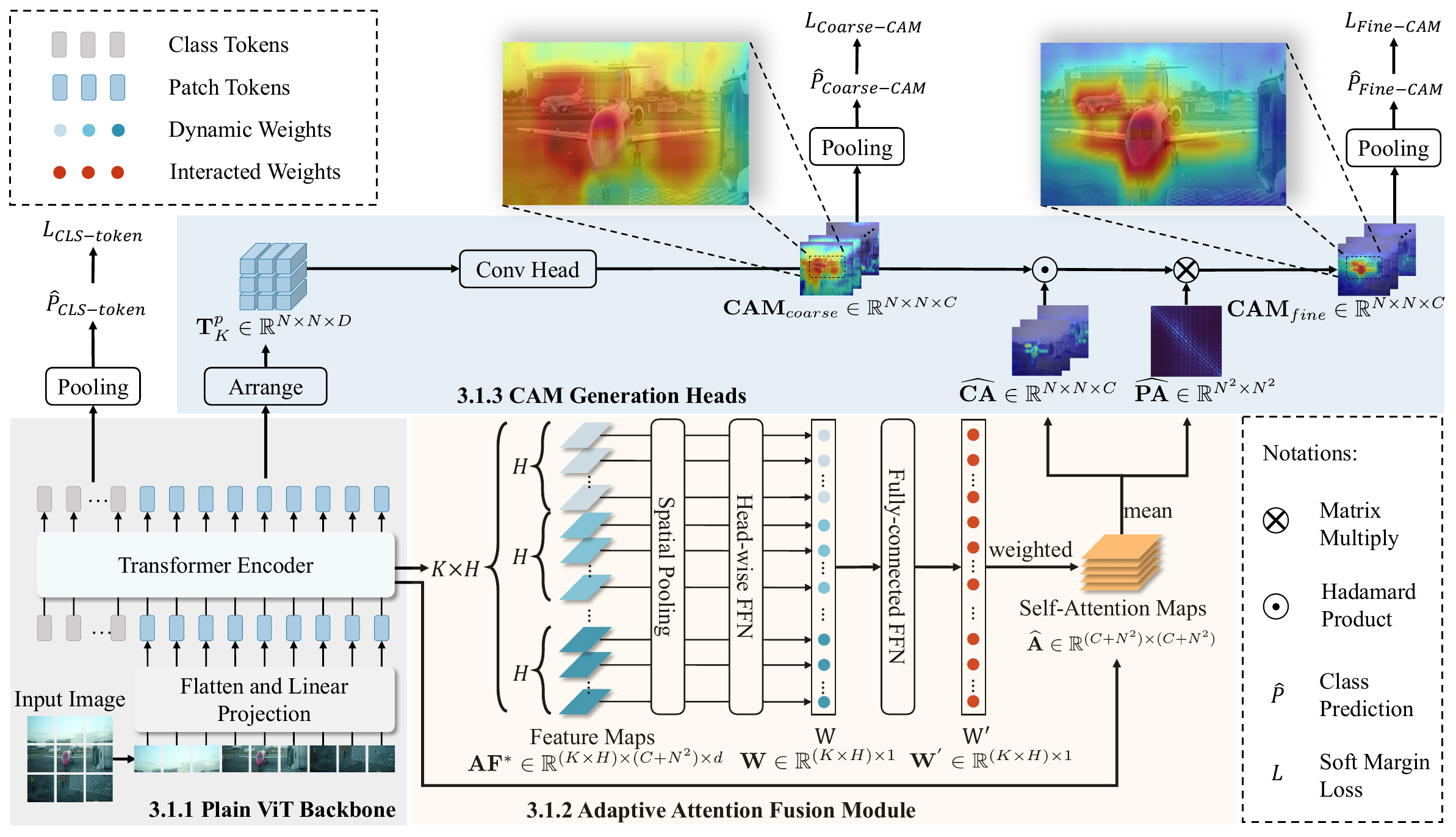}
  
    \caption{An overview of our proposed end-to-end WeakTr CAM generation. WeakTr first inputs the image patch tokens and multiple class tokens into the transformer encoder. Next, we generate coarse CAM by applying a convolution layer to the patch tokens. Then we use the adaptive attention fusion module to generate dynamic weights from all heads' feature maps and make the dynamic weights interact via the interactive feed-forward network (FFN). Finally, we optimize the coarse CAM into the fine CAM by using weighted cross-attention maps and weighted patch-attention maps. Class tokens, coarse CAM, and fine CAM finally generate predictions by pooling to compute the corresponding prediction loss.}
    \label{fig:weaktr}
\vspace{-0.2cm}
\end{figure*}

As shown in Fig.~\ref{fig:weaktr}, our framework uses plain ViT as the backbone. First, we split an input image into $N^2$ patches, flatten them, and linearly map them into $N^2$ patch tokens. Furthermore, we generate $C$ learnable class tokens, where $C$ represents the total number of classification categories, and concatenate them with patch tokens as the transformer encoder's input $ \mathbf{T_{0}} \in \mathbb{R} ^ {(C + N^2) \times D} $, where $D$ is the dimension of input tokens.

ViT mainly relies on the multi-head self-attention mechanism to capture long-range dependencies. Specifically, we first normalize the input sequence and transform it into a triplet of $\mathbf{Q} \in \mathbb{R} ^ {(C + N^2) \times d} $, $\mathbf{K} \in \mathbb{R} ^ {(C + N^2) \times d} $ and $\mathbf{V} \in \mathbb{R} ^ {(C + N^2) \times d} $, where $d$ is the embedding dimension of attention heads. Then, we can calculate the attention weights $\mathbf{A} \in \mathbb{R} ^ {(C + N^2) \times (C + N^2)}$ and self-attention output feature $\mathbf{AF} \in \mathbb{R} ^ {(C + N^2) \times d}$ for each attention head as follows:

\begin{align}
    \mathbf{A} &= \mathrm{softmax}(\mathbf{Q} \cdot \mathbf{K}^{\intercal} / \sqrt{d}) \\
    \mathbf{AF} &= \mathbf{A} \cdot \mathbf{V}.\
\end{align}
Each attention head performs its own attention calculations and stitches the final results together.

The transformer encoder consists of $K$ encoding layers internally. Each layer consists of two sub-layers: a multi-head self-attention (MSA) mentioned before and a multilayer perceptron (MLP). 
Layer Normalization (LN) is applied before every sub-layer, and residual connections are applied after every sub-layer. In the $k$-th encoding layer, we input tokens $\mathbf{T}_{k-1}$ and receive $\mathbf{T}_{k}$. Through $K$ encoding layers, we get $\mathbf{T}_{K} \in \mathbb{R} ^ {(C + N^2) \times D}$ as the final output.

\subsubsection{Adaptive Attention Fusion Module}

The adaptive attention fusion module aims to provide accurate weights for self-attention maps, and the weighted attention maps can more accurately represent the relationship between patches and categories, patches and patches.
For each attention head, a single self-attention map $\mathbf{A}$ has a shape of $(C + N^2)^2$, allowing us to obtain the cross-attention maps $\mathbf{CA}$ of the $C$ class tokens for the $N^2$ patch tokens and the patch-attention maps $\mathbf{PA}$ of the $N^2$ patch tokens relative to themselves. The cross-attention maps $\mathbf{CA}$ have a shape of $N \times N \times C$, and the patch-attention maps $\mathbf{CA}$ have a shape of $N^2 \times N^2$. Considering that the transformer encoder has $K$ encoding layers, each with $H$ attention heads, we can obtain the self-attention maps as $\mathbf{A^*} \in \mathbb{R} ^ {(K \times H) \times (C + N^2) \times (C + N^2)}$, the cross-attention maps as $\mathbf{CA^*} \in \mathbb{R} ^ {(K \times H) \times N \times N \times C}$ and the patch-attention maps as $\mathbf{PA^*} \in \mathbb{R} ^ {(K \times H) \times N^2 \times N^2}$.

In order to combine the representation of all attention heads in all layers, previous WSSS methods \cite{xu2022mctformer, li2022transcam} have directly averaged the self-attention maps of different heads in the same layer, and then summed them by different layers. We find that this mean-sum approach to the deployment of transformer attention is rudimentary. As shown in Fig.~\ref{fig:statistical_results}, most different attention heads focus on different areas and classes. This indiscriminate approach to the attention heads tends to introduce more interference to the activation map of foreground objects. So we propose to utilize an adaptive attention fusion module to estimate the importance of different attention heads.

As shown in Fig.~\ref{fig:weaktr}, first we get the feature maps $\mathbf{AF^*} \in \mathbb{R} ^ {(K \times H) \times (C + N ^ 2) \times d}$ from each head's outputs, where $d=D/H$ is the dimension of features for each attention head. We use max pooling along the $(C+N^2)$ dimension to obtain an output of shape ($K \times H) \times d$. Subsequently, a head-wise MLP is used to extract the dynamic weights $\mathbf{W}$ of shape $(K \times H) \times 1$. Then we use an interactive FFN network to interact with the information among the dynamic weights as follows:
\begin{align}
    \mathbf{W} &= \mathrm{FFN}_\mathrm{Head-wise}(\mathrm{Pooling}(\mathbf{AF^*})) \\
    \mathbf{W}^{'} &= \mathrm{FFN}_\mathrm{Fully-connected}(\mathbf{W}),
\end{align}
where $\mathrm{Pooling}$ is the global max pooling and $\mathbf{W}^{'} \in \mathbb{R} ^ {(K \times H) \times 1} $ is the interacted weights for the attention heads.
Finally, we weighted attention maps $\mathbf{A}^{*}$ using the interacted weights $\mathbf{W}^{'}$ as follows:

\begin{align}
    \widehat{\mathbf{A}} &= \frac{1}{K H} \sum _{i=1} ^{K \cdot H} \mathbf{W}^{'}_i \cdot \mathbf{A}_i,
\end{align}
where $\widehat{\mathbf{A}} \in \mathbb{R} ^ {(C + N^2) \times (C + N^2)}$ is the weighted attention weights we get from self-attention maps by using the interacted weights $\mathbf{W}^{'}$. We can further get weighted cross-attention maps $\widehat{\mathbf{CA}} \in \mathbb{R} ^ {N \times N \times C}$ and weighted patch-attention maps $\widehat{\mathbf{PA}} \in \mathbb{R} ^ {N^2 \times N^2}$ from $\widehat{\mathbf{A}}$.

\subsubsection{CAM Generation Heads}
As shown in Fig.~\ref{fig:weaktr}, we generate coarse CAM and optimize it by the weighted self-attention maps $\widehat{\mathbf{A}}$. In order to get the coarse CAM first, we need to extract the last $N^2$ patch tokens from $\mathbf{T}_{K}$. Then we arrange the $N^2$ patch tokens as $ \mathbf{T}_{K}^p \in \mathbb{R} ^ {N \times N \times D}$. A convolution layer is used to obtain $\mathbf{CAM}_{coarse} \in \mathbb{R} ^ {N \times N \times C}$ as follows:
\begin{align}
    \mathbf{T}_{K}^p &= \mathrm{Arrange}(\mathbf{T}_{K}[C + 1:C + N^2]) \\
    \mathbf{CAM}_{coarse} &= \mathrm{Conv}(\mathbf{T}_{K}^p).
\end{align}

After obtaining the coarse CAM, we refine it using the weighted cross-attention maps $\widehat{\mathbf{CA}}$ and the weighted patch-attention maps $\widehat{\mathbf{PA}}$ extracted from the weighted self-attention maps $\widehat{\mathbf{A}}$. 
We have adopted the same method as MCTformer \cite{xu2022mctformer} and TransCAM \cite{li2022transcam} to combine $\mathbf{CAM}_{coarse}$, $\widehat{\mathbf{CA}}$, and $\widehat{\mathbf{PA}}$:
\begin{align}
    \mathbf{CAM}_{fine} &= \mathfrak{R} ^ {N \times N \times C} (\widehat{\mathbf{PA}} \cdot  \mathfrak{R} ^ {N^2 \times C} (\mathbf{CAM}_{coarse} \odot \widehat{\mathbf{CA}})) ,
\end{align}
where $\mathbf{CAM}_{fine}$ is the CAM guided by $\widehat{\mathbf{CA}}$, and $\widehat{\mathbf{PA}}$, $\mathfrak{R} ^ {N^2 \times C}(\cdot)$ is the operator used to reshape the matrix to $N^2 \times C$, $\mathfrak{R} ^ {N \times N \times C}(\cdot)$ is the operator used to reshape the matrix to $N \times N \times C$, and $\odot$ denotes the Hadamard product. As shown in Table~\ref{tab:ablaaf} and Fig.~\ref{fig:attention_maps}, the weighted self-attention maps provide more accurate guidance for CAM than the mean-sum self-attention maps.

\subsubsection{End-to-End WeakTr Training}

The key to the ability of the adaptive attention fusion module to provide accurate weights lies in our end-to-end training.
In contrast to conventional transformer-based methods, which produce high-quality CAM in the post-process stage, our \name~ end-to-end generates fine CAM in the CAM training stage with the help of the adaptive attention fusion module. Thus we can additionally calculate class prediction and loss corresponding to the fine CAM and optimize the proposed adaptive attention fusion module through image-level supervision. The process of improving coarse CAM using weighted self-attention maps to generate fine CAM and calculating the loss function $L_{Fine-CAM}$ is fully differentiable. Therefore, the loss $L_{Fine-CAM}$ for classification can provide weak supervision guidance for the weight allocation of attention maps. Under this guidance, attention heads that match the object of interest in terms of both attended categories and attended regions are encouraged to have greater weights assigned to them, while those that do not match have smaller weights.

Then, we introduce how to calculate all predictions and losses in \name. For class predictions, we obtained $\hat{P}_{CLS-token}$, $\hat{P}_{Coarse-CAM}$, and $\hat{P}_{Fine-CAM}$ from class tokens $\mathbf{T}_{K}^p $, $\mathbf{CAM}_{coarse}$, and $\mathbf{CAM}_{fine}$, respectively through pooling. For all three losses, we choose to use the multi-label soft margin loss computed between the image-level ground-truth labels $y$ and the class predictions $\hat{P}$ as follows:
\begin{align*}
    Loss(\hat{P}, y) = -\frac{1}{C} \sum_{i}^C y_i \log(\frac{1}{1+\exp(-\hat{P_i})}) \\ 
    + (1-y_i)\log(\frac{\exp(-\hat{P_i})}{1+\exp(-\hat{P_i})}),%
\end{align*}
where $C$ is the number of classification categories. When $\hat{P}$ is $\hat{P}_{CLS-token}$, $\hat{P}_{Coarse-CAM}$ and $\hat{P}_{Fine-CAM}$, $Loss(\hat{P}, y)$ corresponds to $L_{CLS-token}$, $L_{Coarse-CAM}$, and $L_{Fine-CAM}$, respectively.
 We add all the losses shown in Fig.~\ref{fig:weaktr} to get the total loss $\mathcal{L}$ as follows:
\begin{align}
    \mathcal{L}  &= L_{CLS-token} + L_{Coarse-CAM} + L_{Fine-CAM}.
\end{align}

\subsection{Online Retraining Framework of \name~}

To better describe the online retraining phase of \name, we first describe the motivation of the proposed gradient clipping decoder and the gradient clipping rule followed by this decoder. Then we introduce the whole process of the online retraining phase, especially the gradient clipping operation in the decoder.

\subsubsection{Motivation of Online Retraining and Gradient Clipping}
Traditionally, the low quality of CAM in WSSS frameworks requires the CAM refinement \cite{ahn2018psa} phase before they could be used for retraining. This process can be tedious and lengthy (refer to Table~\ref{tab:time_mct}). Our proposed online retraining method involves ViT and a gradient clipping decoder. It can directly train a high-performance semantic segmentation model using CAM, bypassing the need for CAM refinement.

\begin{figure}[htp]
  \centering
  \setlength{\abovecaptionskip}{-0.em}
  \setlength{\tabcolsep}{1.6 pt}
  \includegraphics[width=1.\linewidth]{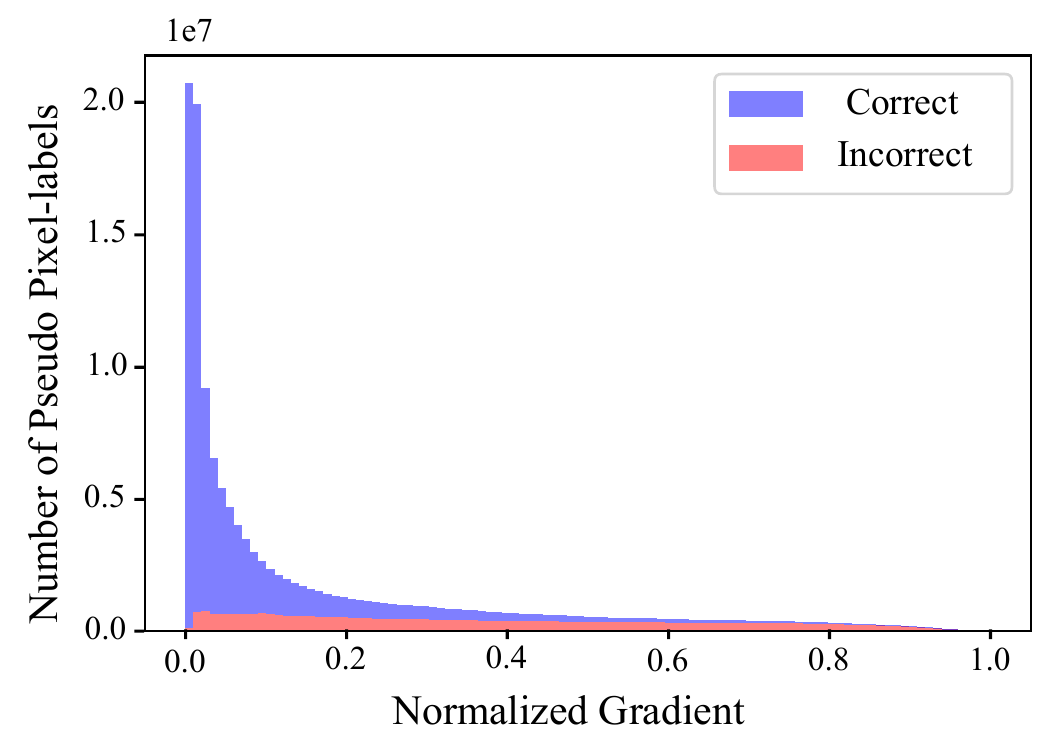}
  \caption{
  The relationship between the number of pseudo pixel-labels and the normalized gradient. The results are obtained from training the Segmenter using pseudo labels in the preliminary training stage.}
   \label{fig:e-pdf}
\end{figure}

As shown in Fig.~\ref{fig:e-pdf}, pixels with smaller gradients are associated with more accurate pseudo labels. Through the proposed gradient clipping, we make the model focus on learning regions that are more likely to be truly annotated.
This is achieved by treating each pixel as a sample and determining whether to clip the gradient at that pixel based on a threshold. The purpose of our gradient clipping decoder is to find a suitable threshold. To determine the gradient threshold, we considered two factors: the gradient across the entire image and the gradient within local regions. Firstly, we use the average gradient value of all pixels in the whole image as the global gradient constraint. Secondly, we divide pixels into patches to obtain local gradient constraints from patch regions, similar to ViT. 
The gradient clipping decoder takes both of these constraints into account when determining whether to clip the gradient of each pixel.
It clips regions with larger gradients, allowing the segmentation network to focus on learning regions with smaller gradients.

\subsubsection{Online Retraining with Gradient Clipping}

\begin{figure}[t]
    \centering
    \includegraphics[width=1.0\linewidth]{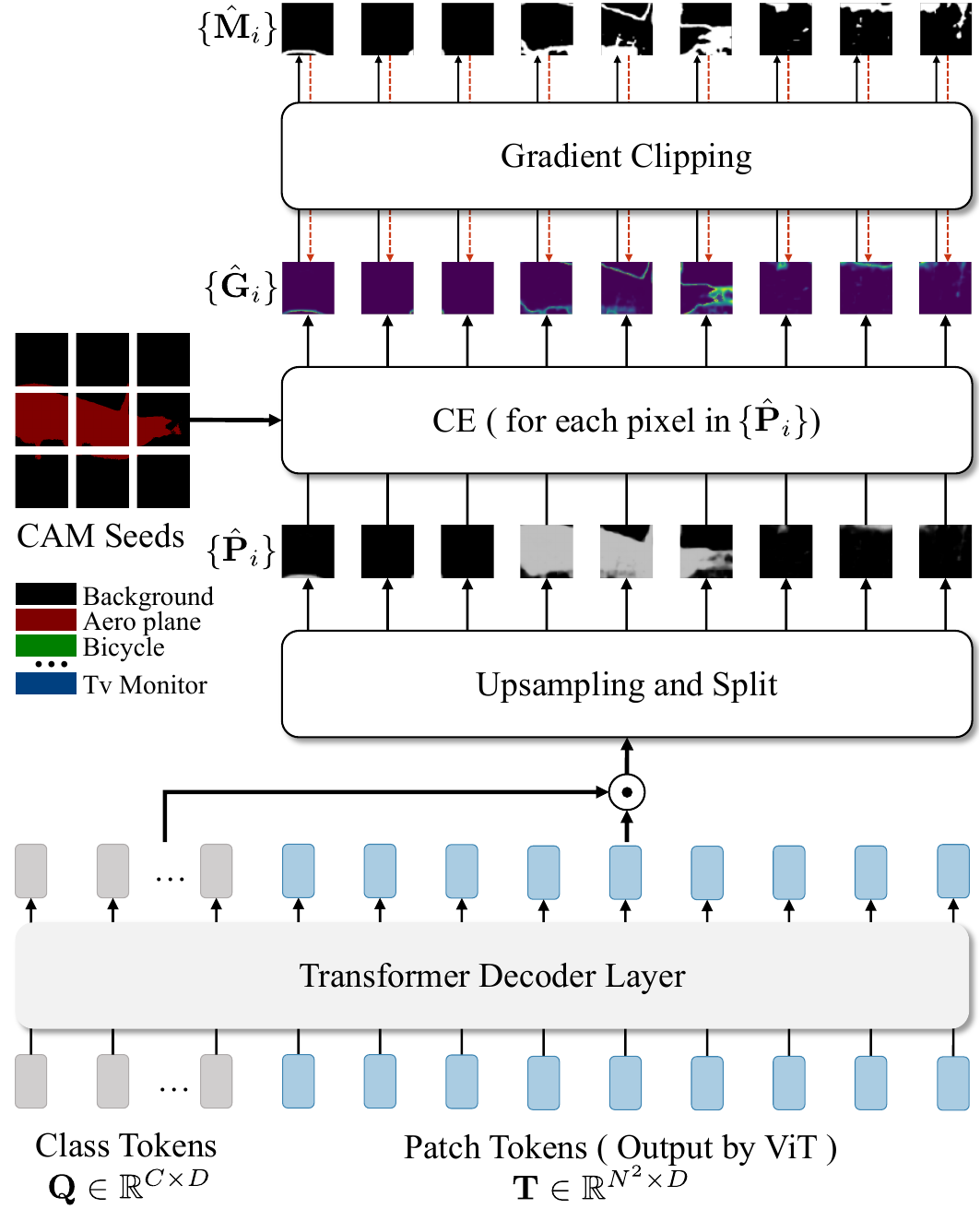}
  
    \caption{Architecture of our proposed gradient clipping decoder. The input of the gradient clipping decoder consists of two parts: class tokens $Q$ and patch tokens $T$ output by the ViT encoder. After the operation of the decoder layers, we obtain prediction patches $\{\hat{\mathbf{P}}_i\}$, gradient patches  $\{\hat{\mathbf{G}}_i\}$, and the gradient clipping mask $\{\hat{\mathbf{M}}_i\}$. }
    \label{fig:gcd}
\vspace{-0.5em}
\end{figure}

As shown in Fig.~\ref{fig:gcd}, first we input the class tokens $\mathbf{Q} \in \mathbb{R} ^ {C \times D}$ and the patch tokens $\mathbf{T} \in \mathbb{R} ^ {N^2 \times D}$ produced by the ViT encoder together into the transformer decoder layer to get the corresponding outputs $\hat{\mathbf{Q}} \in \mathbb{R} ^ {C \times D}$ and $\hat{\mathbf{T}} \in \mathbb{R} ^ {N^2 \times D}$. Next, we use the L2-normalized $\hat{\mathbf{Q}}_{norm}$ and $\hat{\mathbf{T}}_{norm}$ to generate the corresponding predicted sequences. %
Then we pass the predicted sequences through LN and upsample them to get the prediction $\hat{\mathbf{P}} \in \mathbb{R} ^ {O \times O \times C}$ as follows:
\begin{align}
    \hat{\mathbf{P}} &= \mathrm{Upsampling}(\mathrm{LN}(\frac{\hat{\mathbf{T}}_{norm}\cdot \hat{\mathbf{Q}}_{norm}^{\intercal}} {\sqrt{D}})),
\end{align}
where $(O,O)$ is the original resolution of the input image.

To compute the local gradient constraint, we split the prediction $\hat{\mathbf{P}} \in \mathbb{R} ^ {O \times O \times C}$ into $L^2$ non-overlapping patches $\{\hat{\mathbf{P}}_i\}, i \in \{1,\dots,L^2\}$. The shape of each patch in $\{\hat{P}_i\}$ is $S \times S \times C$, and $L = O / S$. Please note that the patch size in $\{\hat{\mathbf{P}}_i\}$ is unrelated to the image patch size in the ViT encoder.
Using the CAM seeds generated from $\mathbf{CAM}_{fine}$ and the prediction patches $\{\hat{\mathbf{P}}_i\}$, we can calculate the gradient patches $\{\hat{\mathbf{G}}_i\}$. Each gradient patch in $\{\hat{\mathbf{G}}_i\}$ has a size of $S \times S$. Meanwhile, we can calculate the average gradient $\{\lambda_i\}$ for each gradient patch in $\{\hat{\mathbf{G}}_i\}$ as follows:
\begin{align}
    \hat{\mathbf{G}}_i &= \mathrm{CE}(\hat{\mathbf{P}}_i, {\mathbf{CAM seeds}}_i), i \in \{1,\dots,L^2\} \\
    \lambda_i & ={\mathrm{mean}(\hat{\mathbf{G}}_i)}, i \in \{1,\dots,L^2\},
\end{align}
where $\mathrm{CE}$ is the cross-entropy loss calculated for each pixel. For each gradient patch in $\{\hat{\mathbf{G}}_i\}$, we use $\{\lambda_i\}$ as a local constraint and the average of $\{\lambda_i\}$ as a global constraint $\lambda_{global}$ as follows: 
\begin{align}
    \lambda_{global} &= \frac{1}{L^2} \sum _{i=1} ^{L^2} \lambda_i.
\end{align}

We choose the maximum of $\lambda_{global}$ and $\{\lambda_i\}$ as the threshold value for clipping mask $\{\hat{\mathbf{M}}_i\}$ generation. In this way, the obtained $\{\hat{\mathbf{M}}_i\}$ considers both local and global gradient constraints, achieving the discarding of patch regions with relatively large gradients.
\begin{align}
    \hat{M}_{i, (j, k)} &=
  \left
  \{\begin{array}{ll}
  1, &  \hat{G}_{i, (j, k)} \le \max (\lambda_i, \lambda_{global})\\
  0, &  \hat{G}_{i, (j, k)}~\textgreater~\max (\lambda_i, \lambda_{global})
  \end{array} 
  \right. ,
\end{align}
where $1 \leq j \leq S$, $1 \leq k \leq S$ and $\mathrm{max}$ is the maximum operation. 

However, the selection of confident CAM regions by the gradient clipping decoder is not reliable enough at the beginning of the segmentation network training. So we set the clipping start value $\tau$ to determine whether to clip. Only when the global mean gradient $\lambda_{global}$ of the current batch is lower than $\tau$, we clip the gradient as follows:

\begin{align}
    \hat{\mathbf{G}}^{'}_i &=
  \left
  \{\begin{array}{ll}
  \hat{\mathbf{G}}_i \odot \hat{\mathbf{M}}_i, &  \lambda_{global} \le \tau \\
  \hat{\mathbf{G}}_i, &  \lambda_{global}~\textgreater~\tau
  \end{array} 
  \right. .
\end{align}

Finally, we get the masked gradient patches $\{\hat{\mathbf{G}}^{'}_i\}$ and back-propagate their average value. By doing so, we dynamically select regions with smaller gradients as confident CAM regions to prioritize learning for the segmentation network. Please note that we only show the structure of the gradient clipping decoder in Fig.~\ref{fig:gcd}. During training, the ViT encoder and gradient clipping decoder are updated together. During inference, we apply the Conditional Random Field (CRF)~\cite{krahenbuhl2011dcrf} to $\{\hat{\mathbf{P}}_i\}$ for improving the segmentation quality.

\begin{table}[htp]
  \centering
  \caption{Evaluation of the CAM pseudo labels in terms of mIoU (\%) on the PASCAL VOC 2012 $train$ and $val$ sets. $\dagger$ indicates using a pre-trained ViT with ImageNet-21k~\cite{deng2009imagenet21k} or other large-scale datasets. Best results are in bold.}
  \begin{tabular}{p{90pt} p{55pt}<{\centering} p{25pt}<{\centering} p{25pt}<{\centering}}
  \toprule
  Method    & Backbone & $train$  & $val$ \\
  \midrule
  \multicolumn{4}{l}{\textbf{\textit{CNN-based methods.}}} \\
  BES\tiny $\ _{\mathrm{ECCV20}}$ \normalsize \cite{chen2020besweakly} & ResNet50 & 67.2 & - \\
  SC-CAM\tiny $\ _{\mathrm{CVPR20}}$ \normalsize \cite{chang2020sccamweakly} & ResNet38   & 63.4 & - \\
  SEAM\tiny $\ _{\mathrm{CVPR20}}$ \normalsize \cite{wang2020seam}    & ResNet38 & 63.6 & - \\
  CONTA\tiny $\ _{\mathrm{NeurIPS20}}$ \normalsize \cite{zhang2020causal}  & ResNet38 & 67.9 & - \\

  AdvCAM\tiny $\ _{\mathrm{CVPR21}}$ \normalsize \cite{lee2021anti} & ResNet50  & 69.9 & -   \\
  ECS-Net\tiny $\ _{\mathrm{ICCV21}}$ \normalsize \cite{sun2021ecs} & ResNet38  & 67.8 & - \\
  OC-CSE\tiny $\ _{\mathrm{ICCV21}}$ \normalsize \cite{kweon2021occse} & ResNet38  & 66.9 & - \\
  VWE\tiny $\ _{\mathrm{IJCV22}}$ \normalsize \cite{ru2022vweweakly} & ResNet101     & 71.4 & - \\
  CLIMS\tiny $\ _{\mathrm{CVPR22}}$ \normalsize \cite{xie2022clims} & ResNet50   & 70.5 & - \\
  Yoon et al.\tiny $\ _{\mathrm{ECCV22}}$ \normalsize \cite{yoon2022aeft} & ResNet38 & 71.0 & - \\
  SFC\tiny $\ _{\mathrm{AAAI24}}$ \normalsize \cite{wu2025pcc} & ResNet50 & 73.7 & - \\
  KTSE\tiny $\ _{\mathrm{ECCV24}}$ \normalsize \cite{chen2024ktse} & ResNet38 & 73.8 & - \\
  \midrule
  \multicolumn{4}{l}{\textbf{\textit{Transformer-based methods.}}} \\
  ViT-PCM$^{\dagger}$\tiny $\ _{\mathrm{ECCV22}}$ \normalsize \cite{rossetti2022vitpcm} & ViT-B & 67.7 & 66.0 \\ %
  AFA\tiny $\ _{\mathrm{CVPR22}}$ \normalsize \cite{ru2022afa} & MiT-B1 & 68.7 & 66.5 \\ %
  ACR\tiny $\ _{\mathrm{CVPR23}}$ \normalsize \cite{kweon2023acr} & DeiT-S & 70.9 & - \\
  ToCo\tiny $\ _{\mathrm{CVPR23}}$ \normalsize \cite{ru2023toco} & DeiT-B & 72.2 & 70.5 \\ %
  ToCo$\dagger$\tiny $\ _{\mathrm{CVPR23}}$ \normalsize \cite{ru2023toco} & ViT-B & 73.6 & 72.3 \\ %
  CLIP-ES$^\dagger$\tiny $\ _{\mathrm{CVPR23}}$ \normalsize \cite{lin2023clipes} & CLIP-ViT-B & 75.0 & - \\ %
  DuPL\tiny $\ _{\mathrm{CVPR24}}$ \normalsize \cite{wu2024dupl} & DeiT-B & 75.1 & 73.5 \\ %
  DuPL$^{\dagger}$\tiny $\ _{\mathrm{CVPR24}}$ \normalsize \cite{wu2024dupl} & ViT-B & 76.0 & 74.1 \\ %
  CTI\tiny $\ _{\mathrm{CVPR24}}$ \normalsize \cite{yoon2024cti} & DeiT-S & 73.7 & - \\ %
  DIAL\tiny $\ _{\mathrm{ECCV24}}$ \normalsize \cite{jang2024dial} & DeiT-S & 71.9 & 70.2 \\ %
  DIAL$^\dagger$\tiny $\ _{\mathrm{ECCV24}}$ \normalsize \cite{jang2024dial} & ViT-S & 75.2 & 73.1 \\ %
  PCSS\tiny $\ _{\mathrm{ECCV24}}$ \normalsize \cite{kwon2024pcss} & DeiT-S & 73.2 & - \\ %
  DiG\tiny $\ _{\mathrm{ECCV24}}$ \normalsize \cite{yoon2024dig} & DeiT-S & 74.3 & - \\ %
  PCC$^\dagger$\tiny $\ _{\mathrm{CVPR25}}$ \normalsize \cite{wu2025pcc} & ViT-B & 74.8 & - \\
  PCRE$^{\dagger}$\tiny $\ _{\mathrm{CVPR25}}$ \normalsize \cite{xu2025pcre} & ViT-B & 77.6 & 76.3 \\ %
  POT$^\dagger$\tiny $\ _{\mathrm{CVPR25}}$ \normalsize \cite{wang2025pot} & CLIP-ViT-B & 79.3 & - \\
  MuP-VSS\tiny $\ _{\mathrm{CVPR25}}$ \normalsize \cite{du2025mupvss} & DeiT-S & 74.1 & - \\
  FFR$^\dagger$\tiny $\ _{\mathrm{CVPR25}}$ \normalsize \cite{yang2025ffr} & ViT-B & - & 76.4 \\
  \midrule
  \multicolumn{4}{l}{\textbf{\textit{Our baseline method.}}} \\
  MCTformer\tiny $\ _{\mathrm{CVPR22}}$ \normalsize \cite{xu2022mctformer} & DeiT-S & 69.1 & - \\
  \midrule
  \multicolumn{4}{l}{\textbf{\textit{Ours.}}} \\
  WeakTr & DeiT-S  & 76.5 & 74.2 \\
  WeakTr$^\dagger$ & DINOv2-S  & 78.1 & 75.6 \\
  WeakTr$^\dagger$ & ViT-S  & \textbf{80.3} & 78.0 \\
  WeakTr$^\dagger$ & EVA-02-S  & 80.0 & \textbf{78.4} \\
  \bottomrule
  \end{tabular}
  \label{tab:camsota}
  \vspace{-0.1 in}
\end{table}

\section{Experiments}
\label{sec: Experiments}
\subsection{Experimental Setup}
\subsubsection{Datasets}
We evaluate our WeakTr on the PASCAL VOC 2012~\cite{everingham2010voc12} dataset and the COCO 2014~\cite{lin2014coco14} dataset. 
The PASCAL VOC 2012 dataset has 20 foreground categories and 1 background category. 
This dataset has three separate splits: the training set (includes 1464 images), the validation set (includes 1449 images), and the test set (includes 1456 images). 
In addition, WSSS methods usually use SBD~\cite{hariharan2011voc12sbd} annotations to increase the training set to 10582 images. 
For another COCO14 dataset, which has 80 object categories for segmentation. The validation set has 40137 images, and the training set has 82081 images. 
We use mean intersection over union (mIoU) to evaluate the validation set in our experiments.

\subsubsection{Implementation Details}
By default, we utilize DeiT-S \cite{touvron2021deit} as the backbone, and all models are trained using the AdamW optimizer \cite{loshchilov2017adamw} to generate CAM. 
We adopt Segmenter \cite{strudel2021segmenter} as the retraining baseline.
During WeakTr's online retraining, we replace Segmenter's decoder with the proposed gradient clipping decoder. 
Following the approach of Segmenter, our online retraining leverages the ViT with AugReg training \cite{steiner2021improvevit}, which is pre-trained on ImageNet-21k \cite{deng2009imagenet21k}
To ensure a fair comparison, we also evaluate our gradient clipping decoder with DeiT-S, which is pre-trained solely on ImageNet-1k. 
Furthermore, to validate the effective utilization of WeakTr for large-scale pre-trained encoders, we assess our gradient clipping decoder with self-supervised pre-trained ViT, DINOv2-S \cite{oquab2023dinov2}, and text-supervised pre-trained ViT, EVA-02-S \cite{fang2023eva}, both of which are pre-trained on large-scale datasets. 
Please note that the gradient clipping decoder is randomly initialized in all experiments.
More training hyper parameters can be found in the appendix.

\subsection{Comparisons with State-of-the-art Methods}
\subsubsection{PASCAL VOC 2012}
At first, we present the quantitative results of CAM pseudo labels for VOC 2012 in Table~\ref{tab:camsota}. 
Previous methods typically generate masks through the CAM refinement phase (e.g., AffinityNet~\cite{ahn2018psa}) and then utilize them as supervision during the retraining phase. 
In contrast, our proposed WeakTr directly employs CRF-processed CAM to provide supervision for the online retraining phase.
Furthermore, the online retraining model also serves to produce both final segmentation results on the $val$ set and masks on the $train$ set. 
It can be observed that the proposed \name{} family achieve superior performance on CAM pseudo masks on both $train$ and $val$ sets.

\begin{table}[t]
  \centering
  \caption{Evaluation of the final segmentation results in terms of mIoU (\%) on the PASCAL VOC 2012 $val$ and $test$ sets. The $Sup.$ column denotes the type of supervision used for training including full supervision ($\mathcal{F}$), image-level labels ($\mathcal{I}$), saliency maps ($\mathcal{S}$), bounding box labels ($\mathcal{B}$), and language supervision ($\mathcal{L}$). $\dagger$ indicates using a pre-trained ViT with ImageNet-21k~\cite{deng2009imagenet21k} or other large-scale datasets. Best results are in bold.}
  
  \begin{tabular}{p{96pt} p{40pt} p{20pt} p{12pt} p{12pt}}
  \toprule
  Method      & Backbone       & $Sup.$                    & $val$  & $test$ \\
  \midrule
  \multicolumn{5}{l}{\textbf{\textit{Fully-supervised Semantic Segmentation (FSSS) methods.}}} \\
  Segmenter\tiny $\ _{\mathrm{ICCV21}}$ \normalsize \cite{strudel2021segmenter} & DeiT-S & \multirow{4}{*}{$\mathcal{F}$}     &   79.7   &  79.6    \\
  Segmenter$^{\dagger}$ \tiny  $\ _{\mathrm{ICCV21}}$ \normalsize \cite{strudel2021segmenter} &  ViT-S    &                  & 82.6    &  83.1    \\
  DeepLabV2\tiny $\ _{\mathrm{TPAMI17}}$ \normalsize \cite{chen2017dplabv2} &  ResNet101    &                  & 77.7     &  79.7    \\
  WideResNet38\tiny $\ _{\mathrm{PR19}}$ \normalsize \cite{wu2019wider} &  ResNet38    &                  & 80.8     &  82.5    \\
  \midrule
  \multicolumn{5}{l}{\textbf{\textit{WSSS methods with bounding box.}}} \\
  BCM\tiny $\ _{\mathrm{CVPR19}}$ \normalsize \cite{song2019boxdriven} & ResNet101    & \multirow{2}{*}{$\mathcal{I} + \mathcal{B}$} & 70.2 &  -   \\
  BBAM\tiny $\ _{\mathrm{CVPR21}}$ \normalsize \cite{lee2021bbam}  & ResNet101   &                        & 73.7 & 73.7 \\
  \midrule
  \multicolumn{5}{l}{\textbf{\textit{WSSS methods with saliency map.}}} \\
  EPS\tiny $\ _{\mathrm{CVPR21}}$ \normalsize \cite{lee2021railroad} & ResNet101    &    \multirow{2}{*}{$\mathcal{I} + \mathcal{S}$}                    & 71.0 & 71.8 \\
  L2G\tiny $\ _{\mathrm{CVPR22}}$ \normalsize \cite{jiang2022l2g} & ResNet101    &                        & 72.1 & 71.7 \\
  \midrule
  \multicolumn{5}{l}{\textbf{\textit{WSSS methods with language supervision.}}} \\
  CLIMS\tiny $\ _{\mathrm{CVPR22}}$ \normalsize \cite{xie2022clims} & ResNet101  &    \multirow{2}{*}{$\mathcal{I} + \mathcal{L}$}                    & 70.4 & 70.0 \\
  CLIP-ES\tiny $\ _{\mathrm{CVPR23}}$ \normalsize \cite{lin2023clipes} & ResNet101    &                        & 72.2 & 72.8 \\
  DIAL$^\dagger$\tiny $\ _{\mathrm{ECCV24}}$ \normalsize \cite{jang2024dial} & ViT-B    &                        & 74.5 & 74.9 \\
  WeCLIP$^\dagger$\tiny $\ _{\mathrm{CVPR24}}$ \normalsize \cite{zhang2024weclip} & CLIP-ViT-B    &                        & 76.4 & 77.2 \\
  POT\tiny $\ _{\mathrm{CVPR25}}$ \normalsize \cite{wang2025pot} & ResNet101    &                        & 76.1 & 76.7 \\
  PCC$^\dagger$\tiny $\ _{\mathrm{CVPR25}}$ \normalsize \cite{wang2025pot} & ViT-B    &                        & 72.2 & - \\
  \midrule
  \multicolumn{5}{l}{\textbf{\textit{WSSS methods with only image-level labels.}}} \\
  OC-CSE\tiny $\ _{\mathrm{ICCV21}}$ \normalsize \cite{kweon2021occse} & ResNet38  &                        & 68.4 & 68.2 \\
  CPN\tiny $\ _{\mathrm{ICCV21}}$ \normalsize \cite{zhang2021cpn}   & ResNet38 &   & 67.8 & 68.5 \\
  VWE\tiny $\ _{\mathrm{IJCV22}}$ \normalsize \cite{ru2022vweweakly}   & ResNet101  &    & 70.6 & 70.7 \\
  SIPE\tiny $\ _{\mathrm{CVPR22}}$ \normalsize \cite{chen2022sipe} & ResNet101 &  & 68.8 & 69.7 \\
  W-OoD\tiny $\ _{\mathrm{CVPR22}}$ \normalsize \cite{lee2022wood} & ResNet101 & & 70.7 & 70.1 \\
  AMN\tiny $\ _{\mathrm{CVPR22}}$ \normalsize \cite{lee2022wood} & ResNet101 & & 69.5 & 69.6 \\
  ViT-PCM\tiny $\ _{\mathrm{ECCV22}}$ \normalsize \cite{rossetti2022vitpcm} & ResNet101 &  & 70.3 & 70.9 \\
  Yoon et al.\tiny $\ _{\mathrm{ECCV22}}$ \normalsize \cite{yoon2022aeft} & ResNet38 &  & 70.9 & 71.7 \\
  ACR\tiny $\ _{\mathrm{CVPR23}}$ \normalsize \cite{kweon2023acr} & ResNet38 &  & 72.4 & 72.4 \\
  OCR\tiny $\ _{\mathrm{CVPR23}}$ \normalsize \cite{cheng2023ocr} & ResNet38 &  & 72.7 & 72.0 \\
  BECO\tiny $\ _{\mathrm{CVPR23}}$ \normalsize \cite{rong2023beco} & MiT-B2 &  & 73.7 & 73.5 \\
  ToCo$^\dagger$\tiny $\ _{\mathrm{CVPR23}}$ \normalsize \cite{ru2023toco} & ViT-B &  & 71.1 & 72.2 \\
  IACD\tiny $\ _{\mathrm{ICASSP24}}$ \normalsize \cite{wu2024iacd} & ResNet101 &  & 71.4 & - \\
  SFC\tiny $\ _{\mathrm{AAAI24}}$ \normalsize \cite{zhao2024sfc} & ResNet38 &  & 70.2 & 71.4 \\
  SFC\tiny $\ _{\mathrm{AAAI24}}$ \normalsize \cite{zhao2024sfc} & ResNet101 &  & 71.2 & 72.5 \\
  DuPL$^\dagger$\tiny $\ _{\mathrm{CVPR24}}$ \normalsize \cite{wu2024dupl} & ViT-B    &                        & 73.3 & 72.8 \\
  CTI\tiny $\ _{\mathrm{CVPR24}}$ \normalsize \cite{yoon2024cti} & ResNet38    &                        & 74.1 & 73.2 \\
  PCSS\tiny $\ _{\mathrm{ECCV24}}$ \normalsize \cite{kwon2024pcss} & ResNet38    &                        & 73.2 & 73.0 \\
  DiG\tiny $\ _{\mathrm{ECCV24}}$ \normalsize \cite{yoon2024dig} & ResNet38    &                        & 73.9 & 73.7 \\
  KTSE\tiny $\ _{\mathrm{ECCV24}}$ \normalsize \cite{chen2024ktse} & ResNet101    &                        & 73.0 & 72.9 \\
  MuP-VSS\tiny $\ _{\mathrm{CVPR25}}$ \normalsize \cite{du2025mupvss} & ResNet38    &                        & 73.6 & 74.7 \\
  PCRE$^\dagger$\tiny $\ _{\mathrm{CVPR25}}$ \normalsize \cite{xu2025pcre} & ViT-B    &                        & 75.5 & 75.9 \\
  FFR$^\dagger$\tiny $\ _{\mathrm{CVPR25}}$ \normalsize \cite{yang2025ffr} & ViT-B    &                        & 76.0 & 75.5 \\
  \midrule
  \multicolumn{5}{l}{\textbf{\textit{Our baseline method.}}} \\
  MCTformer\tiny $\ _{\mathrm{CVPR22}}$ \normalsize \cite{xu2022mctformer} & ResNet38 &     \multirow{2}{*}{$\mathcal{I}$ }                   & 71.9 & 71.6 \\
  MCTformer + WeakTr$^{\dagger}$ & ViT-S &                        & 74.4 & 74.0 \\
  \midrule
  \multicolumn{5}{l}{\textbf{\textit{Ours.}}} \\
  WeakTr & DeiT-S & \multirow{4}{*}{$\mathcal{I}$ }     & 74.0  &    74.1  \\
  WeakTr$^\dagger$ & DINOv2-S  &                         & 75.8  &    75.7  \\
  WeakTr$^\dagger$ & ViT-S  &                         & 78.4  &    79.0  \\
  WeakTr$^\dagger$ & EVA-02-S  &                         & \textbf{78.5}  &    \textbf{79.4}  \\
  \bottomrule
  \end{tabular}
  \label{tab:segsotavoc}
  \vspace{-0.2 in}
\end{table}

Then, we give the quantitative results of the final segmentation results on the VOC 2012 in Table~\ref{tab:segsotavoc}. 
In order to comprehensively compare with mainstream methods, we take both single-stage and multi-stage SoTA methods into account.
Our WeakTr method results are obtained via online retraining using the CRF-processed CAM, and we show the online retraining results using DeiT-S~\cite{touvron2021deit}, DINOv2-S~\cite{oquab2023dinov2}, ViT-S with AugReg training~\cite{steiner2021improvevit}, and EVA-02-S~\cite{fang2023eva}, respectively. 
Our approach outperforms previous techniques on both the $val$ and $test$ sets. 

We also list the fully-supervised methods Segmenter-DeiT-S~\cite{strudel2021segmenter} and Segmenter-ViT-S~\cite{strudel2021segmenter} as upper bounds for our WeakTr. 
We use $\delta$ to express the performance gap between the weakly-supervised method and the upper bound. The $\delta$ of WeakTr-DeiT-S compared to the upper bound method is -5.7\% and -5.5\% for the $val$ and $test$ sets, respectively.
The $\delta$ of WeakTr-ViT-S compared to the upper bound method is -4.2\% and -4.1\% for $val$ and $test$ sets, respectively. 
In summary, our WeakTr approach proves to be better than other SoTA methods at reducing the performance gap between weakly-supervised and fully-supervised methods.

\subsubsection{COCO 2014}
We present the quantitative results of the final segmentation on COCO 2014 in Table~\ref{tab:segsotacoco}. 
Our WeakTr results are obtained through online retraining with the CRF-processed CAM. 
Specifically, our WeakTr-EVA-02-S achieves 9.1\% higher results on the $val$ set compared to MCTformer~\cite{xu2022mctformer}. 
Furthermore, the proposed \name{} with large-scale pre-trained ViTs shows superior performances.  
These results demonstrate the effectiveness of our \name{} and its ability to improve the segmentation performance using large-scale pre-trained backbones, such as DINOv2, ViT, and EVA-02.

\begin{table}[t]
    \centering
    \caption{Evaluation of the final segmentation results in terms of mIoU (\%) on the COCO 2014 $val$ set. $\dagger$ indicates using a pre-trained ViT with ImageNet-21k~\cite{deng2009imagenet21k} or other large-scale datasets. Best results are in bold.}
    \begin{tabular}{llll}
    \toprule
    Method              & Backbone  & $Sup.$ & $val$  \\
    \midrule
    \multicolumn{4}{l}{\textbf{\textit{WSSS methods with saliency map.}}} \\
    EPS\tiny $\ _{\mathrm{CVPR21}}$ \normalsize \cite{lee2021railroad}         & ResNet101 & \multirow{2}{*}{$\mathcal{I}+\mathcal{S}$} & 35.7 \\
    AuxSegNet\tiny $\ _{\mathrm{ICCV21}}$ \normalsize \cite{xu2021auxsegnet}   & ResNet38  & & 33.9 \\
    \midrule
    \multicolumn{4}{l}{\textbf{\textit{WSSS methods with language supervision.}}} \\
    CLIP-ES\tiny $\ _{\mathrm{CVPR23}}$ \normalsize \cite{lin2023clipes} & ResNet101    &\multirow{5}{*}{$\mathcal{I} + \mathcal{L}$}& 45.4 \\
    DIAL$^\dagger$\tiny $\ _{\mathrm{ECCV24}}$ \normalsize \cite{jang2024dial} & ViT-B    &                        & 44.4 \\
    WeCLIP$^\dagger$\tiny $\ _{\mathrm{CVPR24}}$ \normalsize \cite{zhang2024weclip} & CLIP-ViT-B    &                        & 47.1 \\
    POT\tiny $\ _{\mathrm{CVPR25}}$ \normalsize \cite{wang2025pot} & ResNet101    &                        & 47.9 \\
    \midrule
    \multicolumn{4}{l}{\textbf{\textit{WSSS methods with only image-level labels.}}} \\
    OC-CSE\tiny $\ _{\mathrm{ICCV21}}$ \normalsize \cite{kweon2021occse} & ResNet38  &     & 36.4 \\
    CDA\tiny $\ _{\mathrm{ICCV21}}$ \normalsize \cite{su2021context}         & ResNet38  &     & 33.2 \\
    VWE\tiny $\ _{\mathrm{IJCV22}}$ \normalsize \cite{ru2022vweweakly}         & ResNet101  &     & 36.2 \\
    URN\tiny $\ _{\mathrm{AAAI22}}$ \normalsize \cite{li2022uncertainty}         & Res2Net101  &     & 41.5 \\
    SIPE\tiny $\ _{\mathrm{CVPR22}}$ \normalsize \cite{chen2022sipe} & ResNet38 &   & 43.6 \\
    AMN\tiny $\ _{\mathrm{CVPR22}}$ \normalsize \cite{chen2022sipe} & ResNet101 &   & 44.7 \\
    ViT-PCM\tiny $\ _{\mathrm{ECCV22}}$ \normalsize \cite{rossetti2022vitpcm} & ResNet101 &  & 45.0 \\
    Yoon et al.\tiny $\ _{\mathrm{ECCV22}}$ \normalsize \cite{yoon2022aeft} & ResNet38  &        & 44.8 \\
    ACR\tiny $\ _{\mathrm{CVPR23}}$ \normalsize \cite{kweon2023acr} & ResNet38 &  & 45.3 \\
    OCR\tiny $\ _{\mathrm{CVPR23}}$ \normalsize \cite{cheng2023ocr} & ResNet38 &  & 42.5 \\
    BECO\tiny $\ _{\mathrm{CVPR23}}$ \normalsize \cite{rong2023beco} & ResNet101 &  & 45.1 \\
    ToCo$^\dagger$\tiny $\ _{\mathrm{CVPR23}}$ \normalsize \cite{ru2023toco} & ViT-B &  & 42.3 \\
    SFC\tiny $\ _{\mathrm{AAAI24}}$ \normalsize \cite{zhao2024sfc} & ResNet101 &  & 46.8 \\
    DuPL$^\dagger$\tiny $\ _{\mathrm{CVPR24}}$ \normalsize \cite{wu2024dupl} & ViT-B &  & 44.6 \\
    CTI\tiny $\ _{\mathrm{CVPR24}}$ \normalsize \cite{yoon2024cti} & ResNet101 &  & 45.4 \\
    PCSS\tiny $\ _{\mathrm{ECCV24}}$ \normalsize \cite{kwon2024pcss} & ResNet101 &  & 45.7 \\
    DiG\tiny $\ _{\mathrm{ECCV24}}$ \normalsize \cite{yoon2024dig} & ResNet38 &  & 45.5 \\
    KTSE\tiny $\ _{\mathrm{ECCV24}}$ \normalsize \cite{chen2024ktse} & ResNet101 &  & 45.9 \\
    MuP-VSS\tiny $\ _{\mathrm{CVPR25}}$ \normalsize \cite{du2025mupvss} & ResNet38 &  & 46.6 \\
    PCRE$^\dagger$\tiny $\ _{\mathrm{CVPR25}}$ \normalsize \cite{xu2025pcre} & ViT-B &  & 47.2 \\
    FFR$^\dagger$\tiny $\ _{\mathrm{CVPR25}}$ \normalsize \cite{yang2025ffr} & ViT-B &  & 46.8 \\
    \midrule
    \multicolumn{4}{l}{\textbf{\textit{Our baseline method.}}} \\
    MCTformer\tiny $\ _{\mathrm{CVPR22}}$ \normalsize \cite{xu2022mctformer}    & ResNet38  &     & 42.0 \\
    \midrule
    \multicolumn{4}{l}{\textbf{\textit{Ours.}}} \\
    WeakTr & DeiT-S   &  \multirow{4}{*}{$\mathcal{I}$}   & 46.9 \\
    WeakTr$^{\dagger}$ & DINOv2-S   &   & 48.9 \\
    WeakTr$^{\dagger}$ & ViT-S   &    & 50.3 \\
    WeakTr$^{\dagger}$ & EVA-02-S   &  & \textbf{51.1} \\
    \bottomrule
    \end{tabular}
    \label{tab:segsotacoco}
    \vspace{-0.2 in}
\end{table}

\subsection{Ablation Studies}
\subsubsection{Improvements of Adaptive Attention Fusion}
To further analyze the improvements brought by the proposed adaptive attention fusion, we give the quantitative results of CAM on the VOC 2012 $train$ set in Table~\ref{tab:ablaaf}. 
Here, we use MCTformer as the baseline, which aggregates the self-attention maps using mean-sum. 
With the proposed adaptive attention fusion (AAF), the proposed \name{} improves the CAM quality by 5.5\% mIoU.
Additionally, the proposed \name{} can achieve better CAM quality through Conditional Random Field (CRF)~\cite{krahenbuhl2011dcrf}.

\begin{table}[t]
  \centering
  \caption{Ablation study for the adaptive attention fusion module (AAF) in terms of precision (\%), recall (\%), and mIoU (\%) on the PASCAL VOC 2012 $train$ set. ``mean-sum" means the attention maps of ViT are aggregated using mean-sum. 
  ``w/ CRF" means the adoption of CRF for processing. Best results are in bold.}
  \begin{tabular}{lccc}
  
   \toprule
   
   Method (CAM generation)
   & Precision
   & Recall
   & mIoU 
   \\
   \midrule
   
   Baseline (mean-sum)            & 75.0      & 77.9   & 61.7 \\

   WeakTr (w/ AAF)                                      & 77.0     & 83.8   & 67.2 \\

   WeakTr (w/ AAF \& CRF) & $\textbf{78.9}$      & $\textbf{84.9}$   & $\textbf{69.4}$ \\

   \bottomrule
   
   \end{tabular}
  \label{tab:ablaaf}
  \vspace{-0.1 in}
\end{table}

\subsubsection{The Impact of Components in the Adaptive Attention Fusion Module}

We use the adaptive attention fusion (AAF) module to measure the importance of different attention heads. 
The AAF module consists of a pooling layer, an FFN, and a sigmoid activation function. 
We conduct ablation studies for the pooling layer and FFN to determine the impact of each component in the AAF. 

\begin{table}[t]
    \renewcommand\arraystretch{1.4}
    \centering
    \caption{Ablation study for the hidden dimension of FFN in the adaptive attention fusion module in terms of mIoU (\%) on the PASCAL VOC 2012 $train$ set. We mark the best result in bold.}
    \vspace{0.5em}
    \begin{tabular}{c|cccccc}
    \Xhline{1pt}
    hidden dimension &72& 36 & 18 & 9 & 3 \\
    \Xhline{0.5pt}
    $train$ &64.5 & 65.0&65.5 &\textbf{67.2} & 63.8\\
    \Xhline{1pt}
    \end{tabular}
    \label{tab:aafreduction}
    \vspace{-0.1 in}
\end{table}

As shown in Table~\ref{tab:aafreduction}, increasing the hidden dimension of the FFN does not lead to greater improvement. 
This indicates that we only need a lightweight FFN network to fuse the information from the different attention heads. 
Through this fusing operation, we can obtain relatively accurate attention weights.

\begin{table}[t]
    \renewcommand\arraystretch{1.4}
    \centering
    \caption{Ablation study for the different pooling layer in the adaptive attention fusion module in terms of mIoU (\%) on the PASCAL VOC 2012 $train$ set. We mark the best result in bold.}
    \vspace{0.5em}
    \begin{tabular}{c|ccc}
    \Xhline{1pt}
     & max pooling & average pooling \\
    \Xhline{0.5pt}
    $train$ & \textbf{67.2} & 66.8 \\
    \Xhline{1pt}
    \end{tabular}
    \label{tab:aafpooling}
    \vspace{-0.1 in}
\end{table}

As demonstrated in Table~\ref{tab:aafpooling}, the mIoU using average pooling exhibits a slight degradation of $0.4\%$ in comparison with max pooling. 
This observation suggests a slight influence of the pooling operation choice on the CAM outcome.

\begin{figure}[t]

    \includegraphics[width=1.0\linewidth]{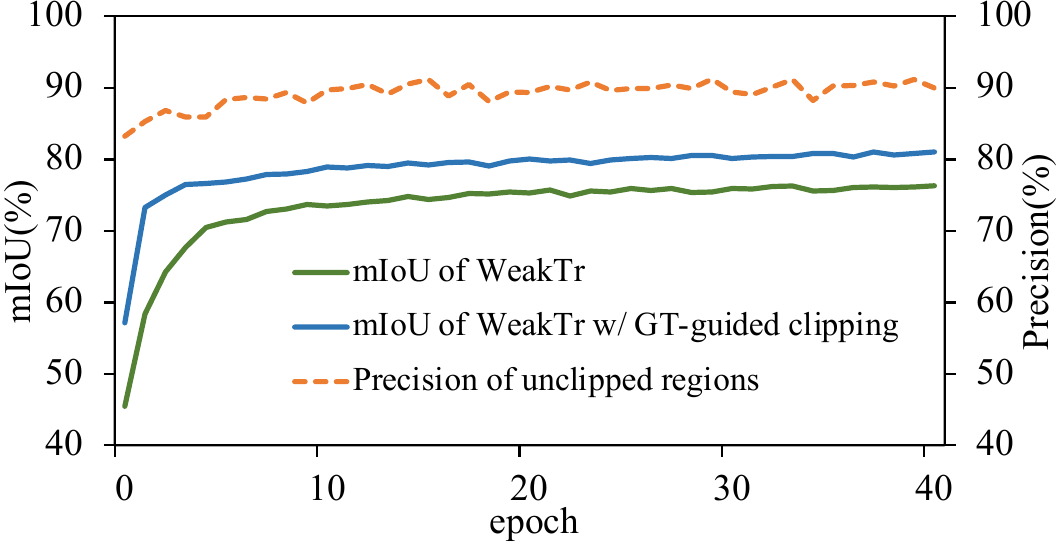}

    \caption{The effectiveness of the gradient clipping decoder and the upper bound for online retraining. `\textbf{orange curve}': the precision of the pseudo labels for unclipped regions in the gradient clipping decoder. `\textbf{green curve}': the mIoU of the WeakTr's online retraining. `\textbf{blue curve}': the mIoU of the WeakTr's online retraining with GT-guided clipping, representing the upper bound of online retraining.}
    \label{fig:WeakTr_mIoU}
\vspace{-0.1 in}
\end{figure}

\begin{table}[htp]
    \centering
    \caption{Ablation study for the start value $\tau$ of the gradient clipping decoder in terms of mIoU (\%) on the PASCAL VOC 2012 $val$ set. We mark the best result in bold.}
    \renewcommand\arraystretch{1.4}

    \begin{tabular}{c|cccccc|c}
    \Xhline{1pt}
    \multirow{3}{*}{\begin{tabular}[c]{@{}c@{}}Naive \\ Decoder\\ (Baseline)\end{tabular}} & \multicolumn{6}{c|}{Gradient Clipping Decoder} & \multirow{3}{*}{$val$} \\ 
    \Xcline{2-7}{0.5pt}
     & \multicolumn{5}{c|}{start value $\tau$} & \multirow{2}{*}{CRF} & \\ 
    \Xcline{2-6}{0.5pt}
     & 1.6 & 1.4 & 1.2 & 1.0 & \multicolumn{1}{c|}{0.8} & & \\
    \Xcline{1-8}{0.5pt}
    \ding{51} & & & & & & & 71.5 \\
     & & &\ding{51}  & & & & 73.0 \\
     & & &\ding{51} & & & \ding{51} & \textbf{74.0} \\
     & \ding{51} & & & & & \ding{51} & 73.6 \\
     & & \ding{51} & & & & \ding{51} & 73.7 \\
     & & &  & \ding{51} & & \ding{51} & 73.5 \\
     & & & & & \ding{51} & \ding{51} & 73.4 \\   
    \Xhline{1pt}
    \end{tabular}
    \label{tab:ablcgd}
    \vspace{-0.15 in}
\end{table}

\subsubsection{Improvements of Gradient Clipping Decoder}
To further analyze the improvements brought by our proposed gradient clipping decoder, we conduct ablation experiments for the gradient clipping decoder and present the results in Table~\ref{tab:ablcgd}. 
Here, we take the naive decoder as our baseline. 
When using a gradient clipping decoder with a start value of 1.2, we could get 1.5\% higher mIoU than the baseline.  
We obtain a 2.5\% higher mIoU than the baseline after processing the results with CRF in the gradient clipping decoder. 
These experiments demonstrate that the proposed gradient clipping decoder is more suitable for the WSSS task than the naive decoder. 

We also conduct an ablation study for the shape of gradient patches, as shown in Table~\ref{tab:ablcgd-patchshape}, a gradient patch with a resolution of (120, 120) can improve the performance of the gradient clipping decoder by allowing it to better utilize the gradient threshold constraint.

\begin{table}[]
    \centering
    \caption{Ablation study for the gradient patches shape $S$ of the gradient clipping decoder in terms of mIoU (\%) on the PASCAL VOC 2012 $val$ set. We mark the best result in bold. }
    \renewcommand\arraystretch{1.4}
    \begin{tabular}{c|ccccc}
    \Xhline{1pt}
    $S$ & 480 & 240 & 160 & 120 & 96\\
    \Xhline{0.5pt}
     $val$ &73.3& 73.3& 73.5& \textbf{74.0}& 73.3 \\   
    \Xhline{1pt}
    \end{tabular}
    \label{tab:ablcgd-patchshape}
    \vspace{-0.2 in}
\end{table}

\begin{table}[htp]
    \centering
    \caption{Ablation study to investigate the impact of utilizing various pre-trained backbones during our online retraining phase. Results are on the PASCAL VOC 2012 and COCO 2014 $val$ sets.}
    \begin{tabular}{lcccc}
        \toprule
        \multirow{2}{*}{\begin{tabular}[c]{@{}l@{}}Pre-trained\\ Encoder\end{tabular}} & \multicolumn{2}{c}{Pre-training Data} & \multicolumn{2}{c}{mIoU on $val$ set} \\
                                                                                     & IN-1k           & Extra          & VOC12   & COCO14   \\
        \midrule
        Random Initialized & \ding{55} & \ding{55} & 11.4 & 9.49 \\ 
        DeiT-S/16~\cite{touvron2021deit} & \ding{51} & \ding{55}  & 74.0       & 46.9        \\
        DINOv2-S/14~\cite{oquab2023dinov2} & \ding{51} & LVD-142M & 75.8       & 48.9        \\
        ViT-S/16~\cite{steiner2021improvevit} & \ding{51} & IN-21k & 78.4       & 50.3        \\
        EVA-02-S/14~\cite{fang2023eva} & \ding{51} & IN-21k, \textbf{EVA}  & 78.5       & 51.1        \\
        \bottomrule
    \end{tabular}
    \label{tab:backbone}
    \vspace{-0.15 in}
\end{table}

We further investigate the effectiveness of the gradient clipping decoder during the training process. 
As shown in Fig.~\ref{fig:WeakTr_mIoU}, the results show that the precision of the regions retained by the gradient clipping decoder is around 90\%, compared to only 78.9\% for CAM in Table~\ref{tab:ablaaf}. 
Although the gradient clipping decoder discards some gradient regions, it ensures that the learned regions are mostly accurate. 
The blue curve also shows the upper bound that online retraining can reach when guided by ground truth for gradient clipping.

\subsubsection{The Impact of Pre-trained ViTs}

Furthermore, we explored the segmentation results obtained by different pre-trained ViTs during the online retraining phase, as shown in Table~\ref{tab:backbone}. 
The different pre-trained ViTs have the same model size and different pre-training data. 
At first, we list the results of random initialized ViT. It only achieves poor results on benchmarks. 
Next, we take DeiT-S/16 as a baseline for pre-trained ViT, which is pre-trained with the IN-1k~\cite{russakovsky2015imagenet1k} dataset. 
The results of DeiT-S/16 significantly outperform the random initialized encoder. 
At last, we show the results of large-scale pre-trained ViTs, which are pre-trained with extra data. 
The DINOv2-S/14 utilizes the LVD-142M dataset~\cite{oquab2023dinov2} as the extra dataset.
The ViT-S/16 is pre-trained with IN-21k~\cite{deng2009imagenet21k} and then fine-tuned with IN-1k. 
The most powerful pre-trained ViT is EVA-02-S/14, which uses \textbf{EVA}-CLIP~\cite{fang2023eva01} 
as the masked image modeling (MIM) teacher and is pre-trained with IN-21k. From the perspective of pre-training data, the ViT backbones pre-trained with large-scale data perform better than the ones only pre-trained with IN-1k (e.g., DeiT-S). 
It proves that our online retraining phase, designed around the ViT framework, effectively harnesses the power of the large-scale pre-trained ViT.

\subsubsection{The Model Complexity of WeakTr's Online Retraining}

After generating the CAM, previous methods typically used the AffinityNet \cite{ahn2018psa} to refine the CAM and then used the segmentation networks for retraining, e.g., WideResNet38 \cite{wu2019wider}. Our proposed online retraining with a gradient clipping decoder, which replaces the CAM refinement and the retraining phases, fully explores the potential of plain ViT. %
At the same input size, we compare the number of parameters and the multiply-add calculations (MACs) for WeakTr's online retraining network, AffinityNet, and WideResNet38.

As shown in Table~\ref{tab:ablparam}, our method has significantly less complexity and parameters than both AffinityNet and WideResNet38. It demonstrates that our online retraining has better performance with lower computational overheads.

\begin{table}[htp]
    \footnotesize
    \centering
    \caption{Complexity of models. AffinityNet is used to refine the CAM for the previous method. WideResNet38 is used to retrain the pseudo masks for the previous method. WeakTr is our online retraining method with the DeiT-S backbone.}
        \vspace{0.5em}

    \begin{tabular}{lccc}
     \toprule
    Model & Image size & \#Params (M) & MACs (G) \\ 
    \midrule
    AffinityNet\tiny $\ _{\mathrm{CVPR18}}$  \scriptsize \cite{ahn2018psa} & 480$\times$480 & 105.3 & 460.2 \\
    WideResNet38\tiny $\ _{\mathrm{PR19}}$  \scriptsize \cite{wu2019wider} & 480$\times$480 & 124.2 & 600.0 \\
    WeakTr (Ours) & 480$\times$480 & 26.3 & 23.0 \\
    \bottomrule
    \end{tabular}
    \label{tab:ablparam}
    \vspace{-.15 in}
\end{table}

\subsubsection{Improvements of Framework Training Time}
To further analyze the improvements in training time brought by WeakTr framework, we conduct experiments to display training time in Table~\ref{tab:time_comp}. 
On the one hand, WeakTr introduces the AAF module for end-to-end training, so the CAM generation phase takes 20 minutes longer than MCTformer. 
On the other hand, WeakTr's online retraining saves more than 2/3 of the time compared to MCTformer's CAM refinement and retraining. 
Overall, the WeakTr takes more than 60\% less time than MCTformer and has a total speed improvement of 2.6 times.

{
\begin{table}[htp]
    \caption{Training time comparisons. We report the detailed training time for MCTformer and our WeakTr. All the experiments were launched using 1 TITAN X GPU.}
    \centering
    \subfloat[\footnotesize{
        Training time of MCTformer framework. 
    MCTformer consists of 3 phases and takes a total of 30.3 hours.}]{
        \begin{tabular}{lccc}
        \toprule
         \multirow{2}{*}{Method}   & CAM   & CAM  & \multirow{2}{*}{Retraining}   \\
         & Generation & Refinement & \\
        \midrule
        MCTformer & 2 hrs 20 mins   &  12 hrs   & 16 hrs \\
        \bottomrule
        \label{tab:time_mct}
        \end{tabular}}
    \hfill
    \subfloat[\footnotesize{
        Training time of WeakTr framework. 
        WeakTr consists of 2 phases and takes a total of 11.6 hours.}]{
        \begin{tabular}{lccc}
        \toprule
        Method   & CAM Generation  &  \multicolumn{2}{c}{Online Retraining}   \\
        \midrule
        WeakTr (Ours) & 2 hrs 40 mins &  \multicolumn{2}{c}{9 hrs} \\
        \bottomrule
        \label{tab:time_weaktr}
        \end{tabular}}
    \label{tab:time_comp}
    \vspace{-0.2 in}
\end{table}
}

\begin{figure*}[htp]
    \centering
    \includegraphics[width=1.0\textwidth]{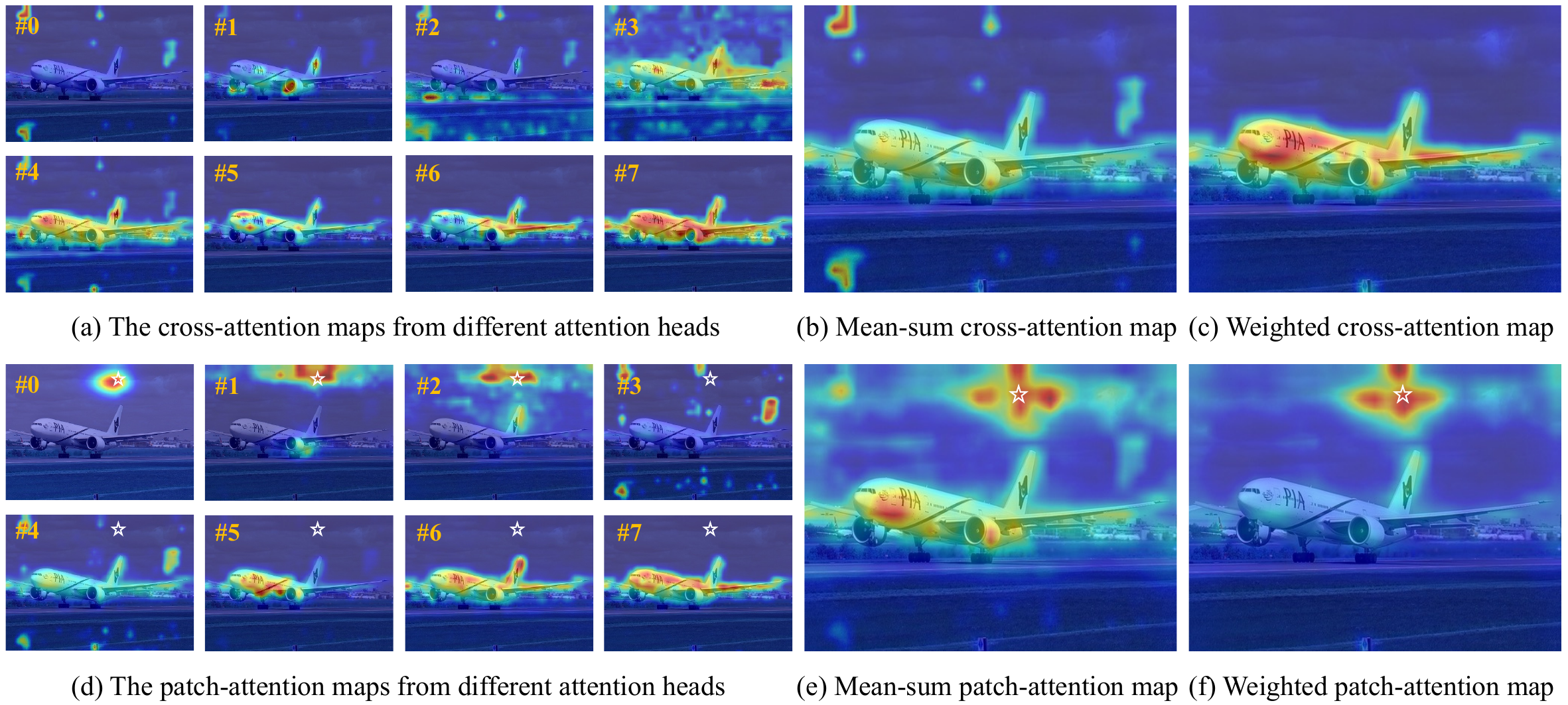}
    \vspace{-.3 in}
    \captionof{figure}{Comparison of the mean-sum results and the weight-based results. (a) shows the cross-attention maps from the different attention heads for the ``plane" category. (b) shows the result obtained by the original mean-sum approach. (c) shows the result obtained by our proposed weight-based approach. (d) shows the patch-attention maps from the different attention heads corresponding to the ``background" point. (We denote the query point with the ``$\star$'') (e) shows the result obtained by the original mean-sum approach. (f) shows the result obtained by our proposed weight-based approach.}
    \label{fig:attention_maps}
    \vspace{-0.1 in}
\end{figure*}

\subsubsection{Improvements of Pure Plain Transformer Framework}
To further analyze the improvements in performance brought by WeakTr, we conduct experiments to display the performance in Table~\ref{tab: abla4baseline}. There is no pure plain Transformer framework previously, while MCTformer is the most related one, but uses \revision{WideResNet38} as the retraining backbone. 
We implement our WeakTr framework on MCTformer.
\revision{When we replace the online retraining with the original CAM refinement and retraining, the mIoU result rises from 71.9 to 73.6.
These results demonstrate the effectiveness of \name{}'s AAF CAM.
}
MCTformer with ViT-S achieves a notable gain but still falls short by 4.0\% mIoU compared to WeakTr, which is a significant margin above the 70\% mIoU. This demonstrates that the proposed WeakTr framework has better performance than the original framework.

\begin{table}[t]
   \centering
   \caption{Comparison with framework performance on VOC12 $val$.}
   \footnotesize
   \setlength{\abovecaptionskip}{0.em}
   \setlength{\belowcaptionskip}{-0.em}
   \setlength{\tabcolsep}{1.6 pt}

   \begin{tabular}{lcccc}%
    \toprule     
     \multirow{2}{*}{Method}
     & CAM
     & Refinement
     & Retraining 
     & \multirow{2}{*}{mIoU} \\
     & Backbone & Backbone & Backbone & 
     \\
     \midrule
     MCTformer original           & DeiT-S & ResNet38 & ResNet38     & $71.9$ \\ %

     \revision{WeakTr + original} & \revision{DeiT-S} & \revision{ResNet38} &  \revision{ResNet38}     & \revision{$73.6$ }\\ %
     
     MCTformer + WeakTr & DeiT-S & - &  ViT-S     & $74.4$ \\ %

     WeakTr  & DeiT-S & - &  ViT-S     & $78.4$ \\ %

     \bottomrule
     \end{tabular}
     
     \label{tab: abla4baseline}
     \vspace{-.25 in}
 \end{table}

\subsection{Visualization}
\subsubsection{The Visual Comparison of Attention Maps}

In Fig.~\ref{fig:attention_maps}, we show the visualization of the cross-attention maps and patch-attention maps. Firstly, as shown in Fig.~\ref{fig:attention_maps} (a-c), we make a comparison between the fused cross-attention map of the ``plane" category obtained by the mean-sum method and our weighted method, respectively. It demonstrates that the mean-sum method is more susceptible to being misled by the incorrect cross-attention maps from different attention heads, as shown in Fig.~\ref{fig:attention_maps} (a). In contrast, our weighted method performs better by avoiding being misled by false information. 

Besides, as shown in Fig.~\ref{fig:attention_maps} (d-f), we also make a comparison between the fused patch-attention map obtained by the two aforementioned methods. 
Specifically, we select to display the patch-attention corresponding to the ``background" query point (denoted with a ``$\star$") and should focus on the ``background" areas. 
However, as shown in Fig.~\ref{fig:attention_maps} (d), there are some patch-attention maps that establish a class activation response with the foreground areas. 
This causes the mean-sum patch-attention map to be misled and creates a connection between the ``background" and ``plane" areas in the final results as shown in Fig.~\ref{fig:attention_maps} (e). 
As shown in Fig.~\ref{fig:attention_maps} (f), our weighted method solved the problem mentioned above correctly.

\subsubsection{More Visualization}
Due to the page limitation, we leave more visualization comparison in the supplementary.

\section{Discussion}
\label{sec: Discussion}

\subsection{Weakly-supervised Semantic Segmentation with Foundation Models}
Recent advancements in foundation models, such as CLIP~\cite{radford2021clip} and the Segment Anything Model (SAM)~\cite{kirillov2023sam}, have significantly contributed to the progress of weakly-supervised semantic segmentation (WSSS).

\paragraph{CLIP}
Trained on 400M image-text pairs, CLIP delivers impressive zero-shot performance in vision-language tasks and has been effectively adapted for WSSS. For example, CLIP-ES~\cite{lin2023clipes} integrates CAM technology with CLIP to track image-level activations as WSSS cues. CLIMS~\cite{xie2022clims} and WeakCLIP~\cite{zhu2024weakclip} leverage CLIP-guided context at the image and pixel levels, respectively. Additionally, WeCLIP~\cite{zhang2024weclip} proposes a single-stage WSSS method using the CLIP-ViT-Base model.

Different from these approaches, our proposed \name{} focuses on analyzing the attention heads of the vision transformer to generate more interpretable class activations. When integrated with ViT-based CLIP models, \name{} can provide deeper insights, which we plan to explore in future work.

\paragraph{Segment Anything Model}
The SAM\cite{kirillov2023sam} demonstrates robust segmentation capabilities, thanks to its large-scale SA-1B dataset\cite{kirillov2023sam}. SA-1B was created through a model-in-the-loop process, enhancing SAM's zero-shot transfer potential. SAM is also designed to be promptable, making it easily combinable with WSSS methods. Building on CLIP-ES, Yang \textit{et al.}\cite{Yang_2024_WACV} merged CLIP and SAM, achieving excellent WSSS performance. SEPL\cite{chen2023sepl} introduces a SAM-supported mask assignment and selection stage to improve pseudo labels, while Sun \textit{et al.}\cite{sun2023altsam} utilizes a SAM-based image grounding method, yielding strong results on WSSS benchmarks. S2C\cite{kweon2024s2c} and WeakSAM~\cite{zhu2024weaksam} propose efficient techniques for extracting accurate masks from CAM.

In summary, integrating powerful foundation models like SAM with WSSS presents a promising research direction. 
Notably, the proposed \name{} also fails to segment objects with complex boundaries, as shown in the visualizations of the supplementary.
In the future, we plan to explore the combination of \name{} and SAM to improve WSSS performance on boundaries further.

\subsection{Weakly-supervised Semantic Segmentation with Advanced Segmenters}

Retraining fully-supervised semantic segmentation methods with pseudo-ground truth is a crucial aspect of WSSS. Earlier approaches often employed DeepLabV1~\cite{liang2015dplabv1} with WideResNet38~\cite{wu2019wider} or DeepLabV2~\cite{chen2017dplabv2}. 
More recent methods have shifted to using more powerful segmenters in the retraining phase. For example, BECO~\cite{rong2023beco} utilizes DeepLabV3+\cite{chen2018dplabv3p} and SegFormer\cite{xie2021segformer} with the MiT-B2 backbone~\cite{xie2021segformer}, while LPCAM~\cite{chen2023LPCAM} and CoSA~\cite{yang2024cosa} integrate the Swin Transformer~\cite{liu2021swin} to enhance performance. 
DHR~\cite{jo2024dhr} further adopts DeepLabV3+ and Mask2Former~\cite{cheng2022mask2former} with the Swin-Large backbone during retraining.

In contrast to these approaches, the proposed \name{} focuses on harnessing the potential of a plain ViT in the WSSS domain.
We use only the plain ViT with a small size and a simple segmenter configuration that consists of a ViT encoder and a segmentation head. 
Our emphasis is on exploring the weakly-supervised learning capabilities of the plain ViT as a generalized model architecture.

\subsection{\revision{Circularity Concern in Gradient Clipping}}

\revision{
Directly using gradients to select confident regions may raise concerns about circularity, and the proposed gradient clipping decoder addresses this concern through the start value $\tau$ of gradient clipping.
Below we clarify how our gradient clipping decoder addresses the circularity concern, and what the potential failure modes are.
}

\revision{
\textbf{How the gradient clipping decoder addresses the circularity concern.}
As the reviewer pointed out, at the beginning of online retraining the overall gradients are typically large, and applying gradient clipping too early may overly weaken the supervision signal, leading to underfitting. 
Therefore, we introduce the clipping start value $\tau$ to explicitly control when gradient clipping is activated (as shown in Eq.~(15) in the main paper).
At the later stage, when the overall gradients become smaller, the clipping mask $\{\hat{\mathbf{M}}_i\}$ simultaneously considers local and global gradient constraints (as shown in Eqs.~(13)--(14) in the main paper), which helps identify relatively unreliable, i.e., less confident, regions even when the global gradient level is low.
To verify that our method can indeed clip unreliable regions and improve the quality of the pseudo ground truth (PGT) at the later stage, we report the precision of the PGT retained by the gradient clipping decoder at different training epochs.
}

\begin{table}[htp]
    \centering
    \caption{\revision{Precision of retained PGT regions on VOC12 $train$ during online retraining with the gradient clipping decoder. The precision steadily improves as training proceeds, indicating that the decoder increasingly discards unreliable pixels.}}
    \begin{tabular}{lcccccc}
        \toprule
        \multirow{2}{*}{\begin{tabular}[c]{@{}l@{}}Settings\end{tabular}} & \multirow{2}{*}{\begin{tabular}[c]{@{}l@{}} PGT\end{tabular}} & \multicolumn{5}{c}{PGT w/ Gradient Clipping Decoder} \\
                                                                                     &           &   1 ep        & 5 ep        & 10 ep        & 20 ep        & 40 ep        \\
        \midrule
        Precision & 78.9 & 82.4 & 87.5 & 88.1 & 89.7 & 91.2 \\ 
        \bottomrule
    \end{tabular}
    \label{tab:pgp_precision_over_epochs}
\end{table}

\revision{
As shown in Table~\ref{tab:pgp_precision_over_epochs}, we also provide quantitative evidence that the retained regions, i.e., unclipped regions, become much cleaner during training: the precision improves steadily from 78.9\%  to 91.2\% after online retraining with the gradient clipping decoder. 
These results demonstrate that the start value $\tau$ avoids ``discarding too many pixels'' in the early stage, and the local-global gradient constraints help clip unreliable pixels at the later stage, thereby steadily improving the label precision of the retained pseudo masks.
}

\revision{
\textbf{Failure modes.}
As discussed above, the start value $\tau$ plays a key role in controlling when gradient clipping is activated. If $\tau$ is set to extreme values, two failure modes may occur:
\begin{itemize}
    \item \textbf{No clipping ($\tau{=}0$):} the model is trained without any gradient clipping and thus is more susceptible to the noise in pseudo ground truth, which makes online retraining degenerate into the standard segmentation model training.
    \item \textbf{Always clipping from the beginning ($\tau$ very large, e.g., $\tau{=}999.9$):} clipping is activated throughout the whole training process. Too many pixels are filtered out early on when gradients are large, which weakens the supervision signal and leads to underfitting.
\end{itemize}
These effects are reflected in our ablation study on $\tau$ as shown in Table~\ref{tab:tau_ablation}, where both ``no clipping'' and ``always clipping'' produce inferior results, while a moderate $\tau$ achieves the best performance. 
We have added this discussion to clarify how $\tau$ balances robustness to pseudo ground truth noise and preserves sufficient supervision for effective learning.
}

\begin{table}[htp]
    \centering
    \caption{\revision{Ablation on the clipping start value $\tau$ in the gradient clipping decoder in terms of mIoU (\%) on VOC12 $val$. $\tau{=}0$ indicates no clipping, while a very large $\tau$ (e.g., 999.9) indicates always clipping from the beginning.}}
    \begin{tabular}{lcccccccc}
        \toprule
        Start Value $\tau$ & 0 & 0.8 & 1.0 & 1.2 & 1.4 & 1.6 & 999.9\\
        \midrule
        mIoU (\%) & 71.8 & 73.4 & 73.5 & 74.0 & 73.7 & 73.6 & 54.2\\ 
        \bottomrule
    \end{tabular}
    \label{tab:tau_ablation}
\end{table}

\revision{
\textbf{Deeper Investigation of Attention Heads.}
While a full theoretical characterization of head specialization in Transformers remains an open problem in the community, we provide additional analyses and visualizations of attention-head specialization and discuss its connection to the pre-training objectives.
}

\begin{figure}[htp]
    \centering
    \includegraphics[width=.48\textwidth]{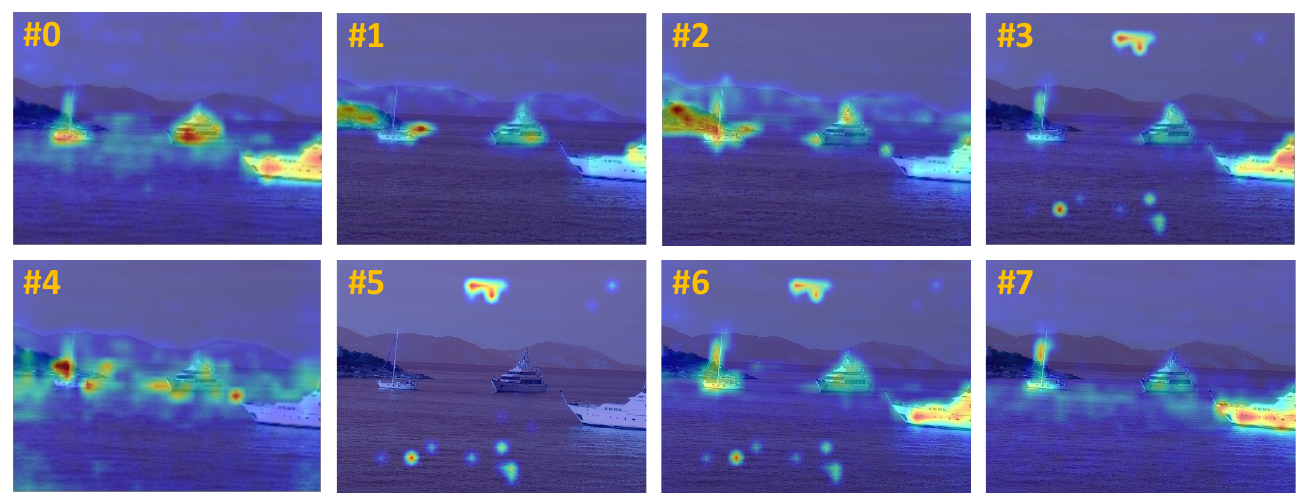}
    \caption{\revision{Attention head specialization for the ``boat'' category. Different attention heads highlight different regions: some aggregate on the boat body, some respond to co-occurring context, i.e., shore, and some focus on a particular spatial region. This motivates using head weighting rather than an unweighted aggregation.}}
    \label{fig:attn_head_boat}
\end{figure}

\revision{
\textbf{Analysis of attention head specialization.} Multi-head attention implements multiple content-based routing functions in parallel, each with its own learned Query($Q$), Key($K$), and Value($V$) projections $(W_Q^h, W_K^h, W_V^h)$, where $h$ is the head index. 
Under the same supervision signal, different heads can minimize the loss by attending to different complementary cues, e.g., objects, regions, or context, which reduces interference and increases representational capacity. 
}

\revision{
\textbf{Evidence from attention-head visualization (``boat'').} We visualize the attention maps of different heads for the boat category in Fig.~\ref{fig:attn_head_boat}. 
As can be seen, different heads attend to different and complementary cues, some heads focus on the boat body/instances, some emphasize co-occurring context such as the shore near the boat, and some exhibit a strong bias toward specific spatial locations, i.e., highlighting a fixed region regardless of the object. 
This diverse behavior directly explains the observed head specialization in Fig.~1 of the main paper.
}

\revision{
\textbf{How this relates to pre-training.} CLIP pre-training encourages patch tokens to organize into semantically meaningful subspaces to support image-text alignment. 
When transferring to WSSS with only image-level supervision, different heads can minimize the loss by routing attention to different predictive cues, i.e., object parts or context, which naturally yields specialization. 
Our method explicitly leverages this diversity by weighting heads that provide consistent category evidence and suppressing heads dominated by context and spatial biases, leading to cleaner localization cues.
}

\revision{
\subsection{Data Scales of ViT Pre-training}
In WSSS, the image-level supervision is coarse and the core difficulty is to recover accurate pixel-level structure from weak signals. Stronger ViT pre-training, which typically leverages more data, improves context modeling ability and transfer robustness. The former helps distinguish target objects from co-occurring background, and the latter reduces overfitting to spurious cues under weak supervision. 
These factors directly translate into higher-quality pseudo masks and stronger retraining performance.
}

\revision{
Consistent with this intuition, Table~\ref{tab:backbone} shows that scaling up the pre-training data consistently improves WSSS performance under a fixed model size, i.e., ViT-S, and the same classification pre-training objective. 
Specifically, without pre-training the model drops to 11.4/9.49 mIoU on VOC/COCO. 
With ImageNet-1K pre-training, i.e., DeiT-S/16, online retraining reaches 74.0/46.9. 
Scaling the pre-training data to ImageNet-21K, i.e., ViT-S/16, further improves to 78.4/50.3, and a stronger IN-21K-based pre-training recipe, i.e., EVA-02-S/14, yields 78.5/51.1. 
In conclusion, stronger pre-trained ViT weights are a plug-in improvement when compute permits, while our framework remains effective under standard IN-1K pre-training.
}

\revision{
\subsection{Gradient Clipping and Hard Examples}
In fully-supervised learning, high-gradient samples often correspond to informative hard examples, e.g., boundaries, and are important for discriminative learning. 
However, \textbf{our setting is weakly supervised}: the gradients are computed, w.r.t. noisy pseudo ground truth, so a large gradient more often indicates strong disagreement with potentially incorrect pseudo ground truth rather than a clean hard example. 
Directly fitting these high-gradient pixels can amplify noise and destabilize online retraining.
}

\revision{
Our gradient clipping decoder does not ``blindly discard boundaries'', instead, it reduces the influence of unreliable regions that dominate the gradient when pseudo ground truth is noisy. 
Empirically, we observe that smaller-gradient pixels are associated with more accurate pseudo ground truth (Fig.~3), and the retained regions, i.e., unclipped regions, exhibit much higher label precision during training (around 90\% in Fig.~5). 
Although some high-gradient regions are clipped, this improves the overall supervision quality and leads to better final segmentation.
}

\revision{
We also avoid overly aggressive clipping at the early stage by using the clipping start value $\tau$ (Eq.~(15)), i.e., clipping is activated only after the model becomes sufficiently stable. As training progresses and pseudo ground truth becomes more consistent, more challenging regions (including boundaries) can be gradually learned instead of being dominated by early noisy gradients.
}

\section{Conclusion}
\label{sec: Conclusion}
We propose \name{} for fully exploring the capacity of plain ViT in the field of weakly-supervised semantic segmentation, achieving superior results of WSSS. The key insights of \name{} are directly generating high-quality CAM in ViT by adaptive multi-layer multi-head attention fusion, and online retraining confident CAM regions with lower gradients through gradient clipping. We hope our work can motivate more studies to understand ViT and propose ViT-based methods to narrow the gap between fully-supervised and weakly-supervised semantic segmentation methods.

\section{Acknowledgements}
\label{sec:ack}
This work was partially supported by the National Natural Science Foundation of China (No. 62276108).

\clearpage

\twocolumn[{%
\renewcommand\twocolumn[1][]{#1}%

\begin{center}
    \centering
    \renewcommand{\thefigure}{A\arabic{figure}}
    \includegraphics[width=1.\textwidth]{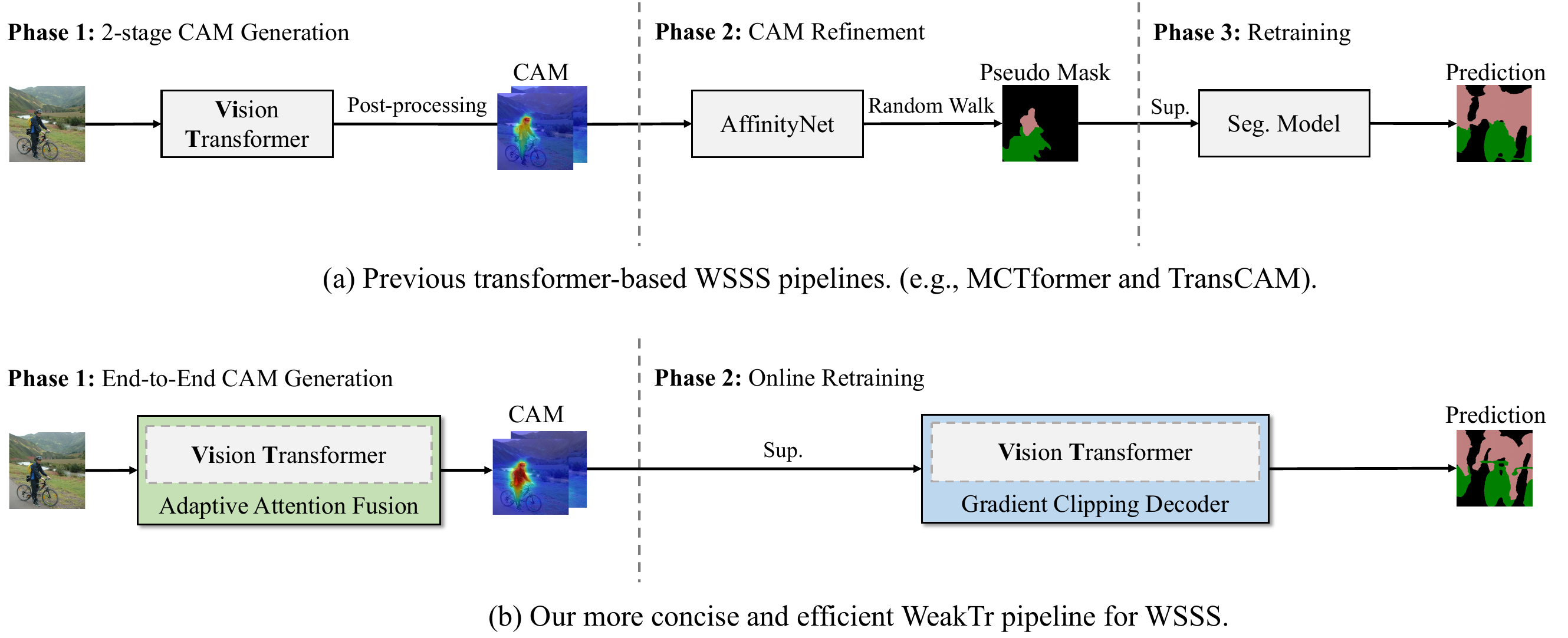}
    \vspace{-0.1 in}
    \captionof{figure}{Comparison of the previous transformer-based WSSS frameworks and our WeakTr framework. Our proposed WeakTr framework is more concise and efficient compared to the previous transformer-based WSSS frameworks. We replace the 2-stage CAM generation with an end-to-end CAM generation using the adaptive attention fusion module, which greatly optimizes the guidance of the transformer attention for the CAM localization. Additionally, we introduce an online retraining method that directly uses CAM for supervised training, eliminating the need for cumbersome CAM refinement and retraining phases. Our gradient clipping decoder enables the network to prioritize learning confident pseudo label regions. The segmentation network of the online retraining method can serve as a segmentation model for inference on the $val$ and $test$ sets and also replace the previous AffinityNet to generate high-quality pseudo masks.}
    \label{fig:frameworks}

\end{center}%
}]

\appendix

\section{}

\subsection{Implementation Details}
\label{sec:supp-imple}

\subsubsection{WeakTr CAM Generation}

During CAM generation, we use the DeiT-S/16 pre-trained on ImageNet as the backbone. The adaptive attention fusion module consists of the global average pooling layer and a 2-layer feed-forward network (FFN) with hidden dimension 18, followed by a sigmoid activation function. During our training process, we use the AdamW \cite{loshchilov2017adamw} optimizer with a batch size of 64 and a weight decay of 0.05. The learning rate is linearly ramped up during the first 5 epochs to its base value, determined with the following linear scaling rule: $lr = 0.0004 \times \mathrm{batchsize}/512$. After the warm-up, we decay the learning rate with a cosine schedule. 
At test time, we use the multi-scale strategy and the CRF \cite{krahenbuhl2011dcrf} for post-processing.

\subsubsection{WeakTr Online Retraining}

At the retraining time, we use the stochastic gradient descent (SGD) \cite{robbins1951sgd} optimizer with a batch size of 4, a momentum parameter of 0.9, and a weight decay of 0. The learning rate is set to 0.0001 and decays using a polynomial scheduler. Besides, we set the hierarchical learning rate for the transformer encoder to be 0.1 times the total learning rate. For the hyperparameters of the gradient clipping decoder, we choose the shape $S$ of 120 for the gradient patches and the start value $\tau$ of 1.2. At test time, we also use the multi-scale strategy and the CRF for post-processing.

\subsection{Additional Ablations}

\label{sec:supp-ablation}

\subsubsection{The Detailed Framework Training Time Comparison}

In Table~\textcolor{red}{7}, we show the framework training time comparison between MCTformer~\cite{xu2022mctformer} and WeakTr. To give a more detailed explanation, we compare the previous transformer-based WSSS frameworks and our WeakTr framework in Fig.~\ref{fig:frameworks} and display the sub-process training time of each phase in Table~\ref{tab:time}. Firstly, we show the MCTformer framework training time in Table~\ref{tab:time} (a). The CAM generation phase consists of network training and post-processing to generate CAM. The CAM refinement phase consists of the affinity label generation from the CAM, AffinityNet~\cite{ahn2018psa} training, and the random walk to refine the CAM. The retraining phase only has a training process. For the comparison, we give the WeakTr framework training time in Table~\ref{tab:time} (b). The CAM generation phase has the aforementioned two processes, which need a little more time because of the AAF module. The online retraining, which replaces the CAM refinement and retraining phases, only has a training process that takes 9 hours, which is 18.8 hours less than the 27.8 hours MCTformer requires for the two phases.
\begin{table*}[]
    \renewcommand{\thetable}{A\arabic{table}}
    \renewcommand\arraystretch{1.4}
    \centering
    \caption{Training time comparisons. We report the more detailed training time for MCTformer and our WeakTr. All the experiments were launched using 1 NVIDIA TITAN X GPU.}
    \subfloat[\footnotesize{Training time of MCTformer framework. MCTformer consists of 3 phases and takes a total of 30.3 hours.}]{
        \begin{tabular}{{m{45pt}| m{50pt}<{\centering} m{60pt}<{\centering} | m{70pt}<{\centering} m{70pt}<{\centering} m{55pt}<{\centering} |m{45pt}<{\centering}}}
            \Xhline{1pt}
             Method   & \multicolumn{2}{c|}{\textbf{CAM Generation}}   & \multicolumn{3}{c|}{\textbf{CAM Refinement}} & \textbf{Retraining}   \\
             \Xcline{0-6}{0.7pt}
             \multirow{2}{*}{MCTformer} & Training & Post-processing & Affinity Label Generation & AffinityNet Training & Random Walk & Training\\
            \Xcline{2-7}{0.7pt}
             & 1 hrs 13 mins & 1 hrs 10 mins & 3 hrs & 8 hrs & 40 mins & 16 hrs \\
            \Xhline{1pt}
        \end{tabular}}
        \vspace{0.3cm}

        \subfloat[\footnotesize{Training time of WeakTr framework. WeakTr consists of 2 phases and takes a total of 11.6 hours.}]{
        \begin{tabular}{{m{45pt}| m{50pt}<{\centering} m{60pt}<{\centering}| m{280pt}<{\centering}}}
            \Xhline{1pt}
             Method   & \multicolumn{2}{c|}{\textbf{CAM Generation}}   & \textbf{Online Retraining}    \\
             \Xcline{0-3}{0.7pt}
             \multirow{2}{*}{WeakTr} & Training & Post-processing & Training  \\
            \Xcline{2-4}{0.7pt}            
            & 1 hrs 23 mins & 1 hrs 20 mins& 9 hrs \\
            \Xhline{1pt}
        \end{tabular}}
        
    \label{tab:time}
\end{table*}

\subsection{Additional Quantitative Results}
\label{sec:results}

\subsubsection{The Gaps between WSSS Methods and Upper Bounds}

As shown in Table~\ref{tab:updelta}, we show the final semantic segmentation results on the PASCAL VOC 2012 $val$ and $test$ sets, as well as the gap $\delta$ between the weakly-supervised method and the upper bound. Among the weakly-supervised semantic segmentation methods based on image-level supervision, our WeakTr achieves a $\delta$ of -5.7\% and -5.5\% for the $val$ and $test$ sets, respectively. For the methods using DeeplabV2 \cite{chen2017dplabv2} pre-trained on COCO as the upper bound, VWE \cite{ru2022vweweakly} obtained the minimum $\delta$ of -7.1\% for the $val$ set and the minimum $\delta$ of -9.0\% for the $test$ set. For the methods using WideResNet38 \cite{wu2019wider} as the upper bound, MCTformer \cite{xu2022mctformer} obtained the minimum $\delta$ of -8.9\% for the $val$ set and Yoon et al. \cite{yoon2022aeft} obtained the minimum $\delta$ of -10.8\% for the $test$ set. Furthermore, our WeakTr$^{\dagger}$ achieves the minimum $\delta$ of -4.2\% and -4.1\% for the $val$ and $test$ sets with the upper bound Segmenter$^{\dagger}$ \cite{strudel2021segmenter} as the upper bound.

These results demonstrate that our proposed online retraining with a gradient clipping decoder takes advantage of the contextual patch tokens output by plain ViT and effectively accomplishes self-correction. Our plain ViT-based online retraining significantly bridges the gap between weakly-supervised and fully-supervised methods, which proves the potential of plain ViT in the WSSS field.

\begin{table}[htp]
    \renewcommand{\thetable}{A\arabic{table}}
    \small
    \renewcommand\arraystretch{1.2}
    \centering
    \caption{Evaluation of the final segmentation results in terms of mIoU (\%) on the PASCAL VOC 2012 $val$ and $test$ sets. The upper bound methods with fully-supervised are denoted by a gray background. The $\dagger$ indicates using the improved ViT pre-trained model. The red numbers denote the performance gaps between the weakly-supervised methods and the upper bounds. We mark the best WSSS results in bold.}
        \vspace{0.5em}

    \begin{tabular}{m{90pt}m{40pt}m{30pt}m{30pt}}
    \toprule
    Method      & Backbone               & $val$ & $test$ \\
    \midrule
    
    \rowcolor{Gainsboro} DeeplabV2\tiny $\ _{\mathrm{TPAMI17}}$  \scriptsize \cite{chen2017dplabv2} &  ResNet101           & 77.7 & 79.7\\
    SC-CAM\tiny $\ _{\mathrm{CVPR20}}$  \scriptsize \cite{chang2020sccamweakly}  & ResNet101  & 66.1$\color{red}_{-11.6}$ & 65.9$\color{red}_{-13.8}$ \\
    VWE\tiny $\ _{\mathrm{IJCV22}}$  \scriptsize \cite{ru2022vweweakly}   & ResNet101  & 70.6$\color{red}_{-7.1}$ & 70.7$\color{red}_{-9.0}$ \\

    CLIMS\tiny $\ _{\mathrm{CVPR22}}$  \scriptsize \cite{xie2022clims} & ResNet101  &     70.4$\color{red}_{-7.3}$ & 70.0$\color{red}_{-9.7}$ \\

    \midrule
    \rowcolor{Gainsboro} WideResNet38\tiny $\ _{\mathrm{PR19}}$  \scriptsize \cite{wu2019wider} &  ResNet38                  & 80.8  & 82.5 \\
    SEAM\tiny $\ _{\mathrm{CVPR20}}$  \scriptsize \cite{wang2020seam}   & ResNet38  & 64.5$\color{red}_{-16.3}$ & 65.7$\color{red}_{-16.8}$ \\
    OC-CSE\tiny $\ _{\mathrm{ICCV21}}$  \scriptsize \cite{kweon2021occse} & ResNet38  & 68.4$\color{red}_{-12.4}$ & 68.2$\color{red}_{-14.3}$ \\
    CPN\tiny $\ _{\mathrm{ICCV21}}$  \scriptsize \cite{zhang2021cpn}& ResNet38 & 67.8$\color{red}_{-13.0}$& 68.5$\color{red}_{-14.0}$ \\
    MCTformer\tiny $\ _{\mathrm{CVPR22}}$  \scriptsize \cite{xu2022mctformer} & ResNet38 & 71.9$\color{red}_{-8.9}$ & 71.6$\color{red}_{-10.9}$ \\
    SIPE\tiny $\ _{\mathrm{CVPR22}}$  \scriptsize \cite{chen2022sipe} & ResNet38 & 68.2$\color{red}_{-12.6}$ & 69.5$\color{red}_{-13.0}$ \\
    W-OoD\tiny $\ _{\mathrm{CVPR22}}$ \scriptsize \cite{lee2022wood} & ResNet38 & 70.7$\color{red}_{-10.1}$ & 70.1$\color{red}_{-12.4}$ \\

    Yoon et al.\tiny $\ _{\mathrm{ECCV22}}$ \scriptsize  \cite{yoon2022aeft}  & ResNet38  & 70.9$\color{red}_{-9.9}$& 71.7$\color{red}_{-10.8}$  \\
    \midrule
    \rowcolor{Gainsboro} Segmenter\tiny $\ _{\mathrm{ICCV21}}$  \scriptsize \cite{strudel2021segmenter} & DeiT-S   &   79.7   & 79.6   \\
        WeakTr (Ours) & DeiT-S    & 74.0$\color{red}_{-5.7}$ & 74.1$\color{red}_{-5.5}$  \\
    \midrule
    \rowcolor{Gainsboro} Segmenter$^{\dagger}$\tiny  $\ _{\mathrm{ICCV21}}$  \scriptsize \cite{strudel2021segmenter}&  ViT-S                & 82.6  & 83.1  \\
    WeakTr$^{\dagger}$ (Ours) & ViT-S                    & \textbf{78.4$\color{red}_{-4.2}$} & \textbf{79.0}$\color{red}_{-4.1}$  \\

    \bottomrule
    \end{tabular}
    \label{tab:updelta}
    \vspace{-0.5em}
\end{table}

\subsubsection{Per-class Semantic Segmentation Results}

\noindent{\textbf{PASCAL VOC 2012.}}
In Table~\ref{tab:valcom} and Table~\ref{tab:testcom}, we compare the per-class segmentation results on the $val$ and $test$ sets for PASCAL VOC 2012. Our WeakTr and WeakTr$^\dagger$ perform better than other state-of-the-art methods, which demonstrates that our plain ViT-based WeakTr can perform well in the WSSS domain.

\noindent{\textbf{COCO 2014.}}
We also give a comparison of the per-class segmentation results on the $val$ set of COCO 2014 in Table~\ref{tab:valcom-coco}. The comparison results show that our WeakTr and WeakTr$^\dagger$ outperform the state-of-the-art methods in most categories, which demonstrates the outstanding performance of our method.

\begin{table*}[t]
    \renewcommand{\thetable}{A\arabic{table}}
    \renewcommand\arraystretch{1.4}
    \small %
    
    \centering
    \caption{Comparison of per-class segmentation results in terms of IoUs on the PASCAL VOC 2012 $val$ set The $^\dagger$ indicates online retraining using the improved ViT pre-trained model. We mark the best results in bold. }
    \begin{tabular}{l|ccccccccccc}
            \Xhline{1pt}
	Method & bkg & plane & bike & bird & boat & bottle & bus & car & cat & chair & cow \\
            \Xhline{0.5pt}
	SEAM \tiny$_\mathrm{CVPR20}$\scriptsize \cite{wang2020seam} & 88.8& 68.5& 33.3& 85.7& 40.4& 67.3& 78.9& 76.3& 81.9& 29.1& 75.5 \\
	AdvCAM \tiny$_\mathrm{CVPR21}$\scriptsize \cite{lee2021anti} & 90.0& 79.8& 34.1& 82.6& 63.3& 70.5& 89.4& 76.0& 87.3& 31.4& 81.3\\
	CPN \tiny$_\mathrm{ICCV21}$\scriptsize \cite{zhang2021cpn} & 89.9& 75.1& 32.9& 87.8& 60.9& 69.5& 87.7& 79.5& 89.0& 28.0& 80.9\\
	OC-CSE \tiny$_\mathrm{ICCV21}$\scriptsize \cite{kweon2021occse} & 90.2 & 82.9 & 35.1 & 86.8 & 59.4 & 70.6 & 82.5 & 78.1 & 87.4 & 30.1 & 79.4 \\
	MCTformer \tiny$_\mathrm{CVPR22}$\scriptsize \cite{xu2022mctformer} & 91.9 &78.3 &39.5 &89.9 &55.9 &76.7 &81.8 &79.0 &90.7 &32.6 &87.1 \\
	WeakTr (Ours) & 92.4 & 88.6 & 44.4 & 89.9 & 71.0 & 80.8 & 88.9 & 80.4 & 93.1 & 35.5 & 85.2 \\
	WeakTr$^\dagger$ (Ours) & \textbf{93.7} & \textbf{90.0} & \textbf{49.9} & \textbf{93.1} & \textbf{76.5} & \textbf{81.8} & \textbf{90.6} & \textbf{86.6} & \textbf{93.6} & \textbf{45.7} & \textbf{93.7} \\
            \Xhline{0.5pt}
	Method & table & dog & horse & mbk & person & plant & sheep & sofa & train & tv & \textbf{mIoU}\\
            \Xhline{0.5pt}
	SEAM \tiny$_\mathrm{CVPR20}$\scriptsize \cite{wang2020seam} & 48.1& 79.9& 73.8& 71.4& 75.2& 48.9& 79.8& 40.9& 58.2& 53.0& 64.5 \\
	AdvCAM \tiny$_\mathrm{CVPR21}$\scriptsize \cite{lee2021anti} & 33.1& 82.5& 80.8& 74.0& 72.9& 50.3& 82.3& 42.2& 74.1& 52.9& 68.1 \\
	CPN \tiny$_\mathrm{ICCV21}$\scriptsize \cite{zhang2021cpn} & 34.8& 83.4& 79.7& 74.7& 66.9& 56.5& 82.7& 44.9& 73.1& 45.7& 67.8\\
	OC-CSE \tiny$_\mathrm{ICCV21}$\scriptsize \cite{kweon2021occse} & 45.9 & 83.1 & 83.4 & 75.7 & 73.4 & 48.1 & 89.3 & 42.7 & 60.4 & 52.3 & 68.4 \\
	MCTformer \tiny$_\mathrm{CVPR22}$\scriptsize \cite{xu2022mctformer} &57.2 &87.0 &84.6 &77.4 &79.2 &55.1 &89.2 &47.2 &70.4 &58.8 &71.9 \\
	WeakTr (Ours) & 50.8 & 85.5 & 84.4 & 78.4 & 76.9 & \textbf{60.0} & 90.2 & 44.0 & 76.6 & 56.2 & 74.0 \\
	WeakTr$^\dagger$ (Ours) & \textbf{57.7} & \textbf{90.5} & \textbf{90.9} & \textbf{81.5} & \textbf{80.9} & 59.6 & \textbf{93.2} & \textbf{58.0} & \textbf{78.1} & \textbf{59.6} & \textbf{78.4} \\
            \Xhline{1pt}
    \end{tabular}
    \label{tab:valcom}
\end{table*}

\begin{table*}[htp]
    \renewcommand{\thetable}{A\arabic{table}}
    \renewcommand\arraystretch{1.4}
    \small %
    \centering
    \caption{Comparison of per-class segmentation results in terms of IoUs on the PASCAL VOC 2012 $test$ set. The $^\dagger$ indicates online retraining using the improved ViT pre-trained model. We mark the best results in bold.}
        \vspace{0.5em}

    \begin{tabular}{l|ccccccccccc}
            \Xhline{1pt}
	Method & bkg & plane & bike & bird & boat & bottle & bus & car & cat & chair & cow \\
            \Xhline{0.5pt}
	AdvCAM \tiny$_\mathrm{CVPR21}$\scriptsize \cite{lee2021anti} & 90.1& 81.2& 33.6& 80.4& 52.4& 66.6& 87.1& 80.5& 87.2& 28.9& 80.1 \\
	CPN \tiny$_\mathrm{ICCV21}$\scriptsize \cite{zhang2021cpn} & 90.4& 79.8& 32.9& 85.8& 52.9& 66.4& 87.2& 81.4& 87.6& 28.2& 79.7\\
	MCTformer \tiny$_\mathrm{CVPR22}$\scriptsize \cite{xu2022mctformer} & 92.3 & 84.4 & 37.2& 82.8& 60.0& 72.8 & 78.0& 79.0& 89.4& 31.7& 84.5\\
	WeakTr (Ours) & 92.7 & \textbf{90.4} & 45.9 & 81.6 & 71.2 & 72.8 & \textbf{90.5} & 82.7 & 92.6 & 31.9 & 77.9 \\
	 WeakTr$^\dagger$ (Ours) & \textbf{94.0} & 89.3 & \textbf{49.3} & \textbf{89.7} & \textbf{72.9} & \textbf{78.3} & 87.9 & \textbf{88.7} & \textbf{95.8} & \textbf{40.0} & \textbf{91.5}\\
            \Xhline{0.5pt}
	
	Method & table & dog & horse & mbk & person & plant & sheep & sofa & train & tv & \textbf{mIoU}\\
            \Xhline{0.5pt}
    
	AdvCAM \tiny$_\mathrm{CVPR21}$\scriptsize \cite{lee2021anti}& 38.5& 84.0& 83.0& 79.5& 71.9& 47.5& 80.8& 59.1& 65.4& 49.7& 68.0 \\

	CPN \tiny$_\mathrm{ICCV21}$\scriptsize \cite{zhang2021cpn}&50.2& 82.9& 80.4& 78.9& 70.6& 51.2& 83.4& 55.4& 68.5& 44.6& 68.5\\

	MCTformer \tiny$_\mathrm{CVPR22}$\scriptsize \cite{xu2022mctformer}& 59.1& 85.3& 83.8& 79.2& \textbf{81.0}& 53.9& 85.3& 60.5& 65.7& 57.7& 71.6 \\

	WeakTr (Ours) & 58.2 & 89.4 & 80.6 & 81.2 & 78.2 & 70.1 & 86.1 & 60.0 & 70.0 & 52.8 & 74.1 \\

    WeakTr$^\dagger$ (Ours) & \textbf{66.3} & \textbf{91.7} & \textbf{91.8} & \textbf{89.2} & 80.7 & \textbf{72.7} & \textbf{92.1} & \textbf{69.3} & \textbf{70.1} & \textbf{57.2} & \textbf{79.0} \\
    
            \Xhline{1pt}
    \end{tabular}
    \label{tab:testcom}
\end{table*}

\begin{table*}[htp]
    \renewcommand{\thetable}{A\arabic{table}}
    \renewcommand\arraystretch{1.4}
    \small %

    \centering
        \vspace{0.5em}

    \caption{Comparison of per-class segmentation results in terms of IoUs on the COCO 2014 $val$ set. We mark the best results in bold.}
    \begin{tabular}{m{60pt} m{45pt}<{\centering} m{45pt}<{\centering} m{45pt}<{\centering} | m{55pt} m{45pt}<{\centering} m{45pt}<{\centering} m{45pt}<{\centering}}
            \Xhline{1pt}
    Class
    & MCTformer \tiny$\mathrm{CVPR22}$\scriptsize \cite{xu2022mctformer}   & WeakTr (Ours)  & WeakTr$^{\dagger}$ (Ours)
    & Class 
    & MCTformer \tiny$\mathrm{CVPR22}$\scriptsize \cite{xu2022mctformer}   & WeakTr (Ours) & WeakTr$^{\dagger}$ (Ours) \\
            \Xhline{0.5pt}
background  &82.4& 82.9& \textbf{84.3}
    &wine glass& 27.0 & 28.4 &\textbf{36.1}\\
    person &62.6& 65.0 & \textbf{67.8}
    &cup& 29.0 & 27.8 &\textbf{42.2}\\
    bicycle&47.4& 51.4& \textbf{53.9}
    &fork & 23.4&24.0&\textbf{28.6}\\
    car& 47.2& 47.2& \textbf{48.8}
    &knife& 12.0& 23.0&\textbf{30.1}\\
    motorcycle&63.7& 66.8& \textbf{69.2}
    &spoon&6.6 &16.5&\textbf{17.0}\\
    airplane&64.7& 69.4& \textbf{72.7}
    &bowl& 22.4&31.7&\textbf{36.8}\\
    bus& 64.5& 64.0& \textbf{65.5}
    &banana &63.2&72.5&\textbf{74.8}\\
    train& 64.5& 65.0& \textbf{71.5}
    &apple &44.4&56.6&\textbf{61.6}\\
    truck & 44.8& 47.9&\textbf{49.1}
    &sandwich& 39.7& 46.8&\textbf{52.7}\\
    boat& 42.3& 47.2&\textbf{47.4}
    &orange& 63.0&70.9&\textbf{72.1}\\
    traffic light& 49.9& 53.7& \textbf{57.0}
    &broccoli& 51.2 &62.5&\textbf{66.4}\\
    fire hydrant&73.2& 76.0&\textbf{76.2}
    &carrot& 40.0&47.1&\textbf{54.2}\\
    stop sign& 76.6& 77.7& \textbf{79.8}
    &hot dog& 53.0&54.7&\textbf{56.9}\\
    parking meter& 64.4& 71.8&\textbf{73.9}
    &pizza& 62.2&74.3&\textbf{81.0}\\
    bench& 32.8& 41.4&\textbf{43.4}
    &donut&  55.7&62.7&\textbf{70.6}\\
    bird& 62.6&67.8&\textbf{70.3}
    &cake& 47.9&55.3&\textbf{62.5}\\
    cat& 78.2&81.5&\textbf{83.5}
    &chair& 22.8&26.5&\textbf{29.2}\\
    dog&68.2&77.0&\textbf{78.8}
    &couch& 35.0&43.8&\textbf{44.5}\\
    horse& 65.8&71.1&\textbf{73.3}
    &potted plant& 13.5&17.7&\textbf{22.3}\\
    sheep& 70.1& 73.4&\textbf{77.7}
    &bed& 48.6&53.3&\textbf{54.8}\\
    cow& 68.3& 70.9&\textbf{77.7}
    &dining table& 12.9&14.7&\textbf{20.5}\\
    elephant& 81.6& 84.1&\textbf{84.4}
    &toilet& 63.1&63.8&\textbf{67.4}\\
    bear& 80.1 &85.2&\textbf{85.5}
    &tv &47.9&53.2&\textbf{54.9}\\
    zebra & \textbf{83.0}& 82.3&81.7
    &laptop& 49.5&46.5&\textbf{52.9}\\
    giraffe& 76.9&\textbf{78.8}&77.7
    &mouse& 13.4&\textbf{11.5}&11.1\\
    backpack& 14.6&20.3&\textbf{22.2}
    &remote& 41.9&43.0&\textbf{47.4}\\
    umbrella& 61.7& 68.2&\textbf{69.8}
    &keyboard& 49.8&52.0&\textbf{55.5}\\
    handbag& 4.5& \textbf{7.2}&7.1
    &cellphone& 54.1&56.2&\textbf{64.1}\\
    tie&25.2& 28.5&\textbf{33.3}
    &microwave& 38.0&40.0&\textbf{50.1}\\
    suitcase& 46.8&52.0&\textbf{59.3}
    &oven& 29.9&36.3&\textbf{39.3}\\
    frisbee& 43.8& 57.8&\textbf{65.0}
    &toaster&0.0&0.0&\textbf{4.9}\\
    skis& 12.8& 15.8&\textbf{16.2}
    &sink& \textbf{28.0}&23.4&19.2\\
    snowboard& 31.4& 36.9&\textbf{40.0}
    &refrigerator& 40.1&52.2&\textbf{53.1}\\
    sports ball& 9.2& \textbf{32.0}&21.2
    &book& 32.2&35.2&\textbf{38.9}\\
    kite& 26.3& 41.4&\textbf{55.3}
    &clock&\textbf{43.2}&41.7&38.1\\
    baseball bat& 0.9& 1.2&\textbf{2.7}
    &vase & 22.6&27.6&\textbf{31.7}\\
    baseball glove& 0.7&0.4&\textbf{5.3}
    &scissors& 32.9&44.2&\textbf{50.9}\\
    skateboard& 7.8&12.8&\textbf{13.1}
    &teddy bear& 61.9&66.4&\textbf{68.2}\\
    surfboard& 46.5&55.4&\textbf{63.3}
    &hair drier& 0.0&\textbf{0.2}&0.0\\
    tennis racket& 1.4& 8.2&\textbf{11.9}
    &toothbrush&12.2&18.9&\textbf{33.8}\\
    \Xcline{5-8}{0.7pt}
    bottle&31.1&38.2&\textbf{42.5}
    &\textbf{mIoU}& 42.0&46.9&\textbf{50.3}\\
            \Xhline{1pt}
    \end{tabular}
    \label{tab:valcom-coco}
    \vspace{-0.5em}
\end{table*}

\subsection{Additional Visualization Results}
\label{sec:visual}

\subsubsection{The CAM and Mask Results}

\begin{figure*}[ht]
    \renewcommand{\thefigure}{A\arabic{figure}}

    \centering
    \includegraphics[width=1.\linewidth]{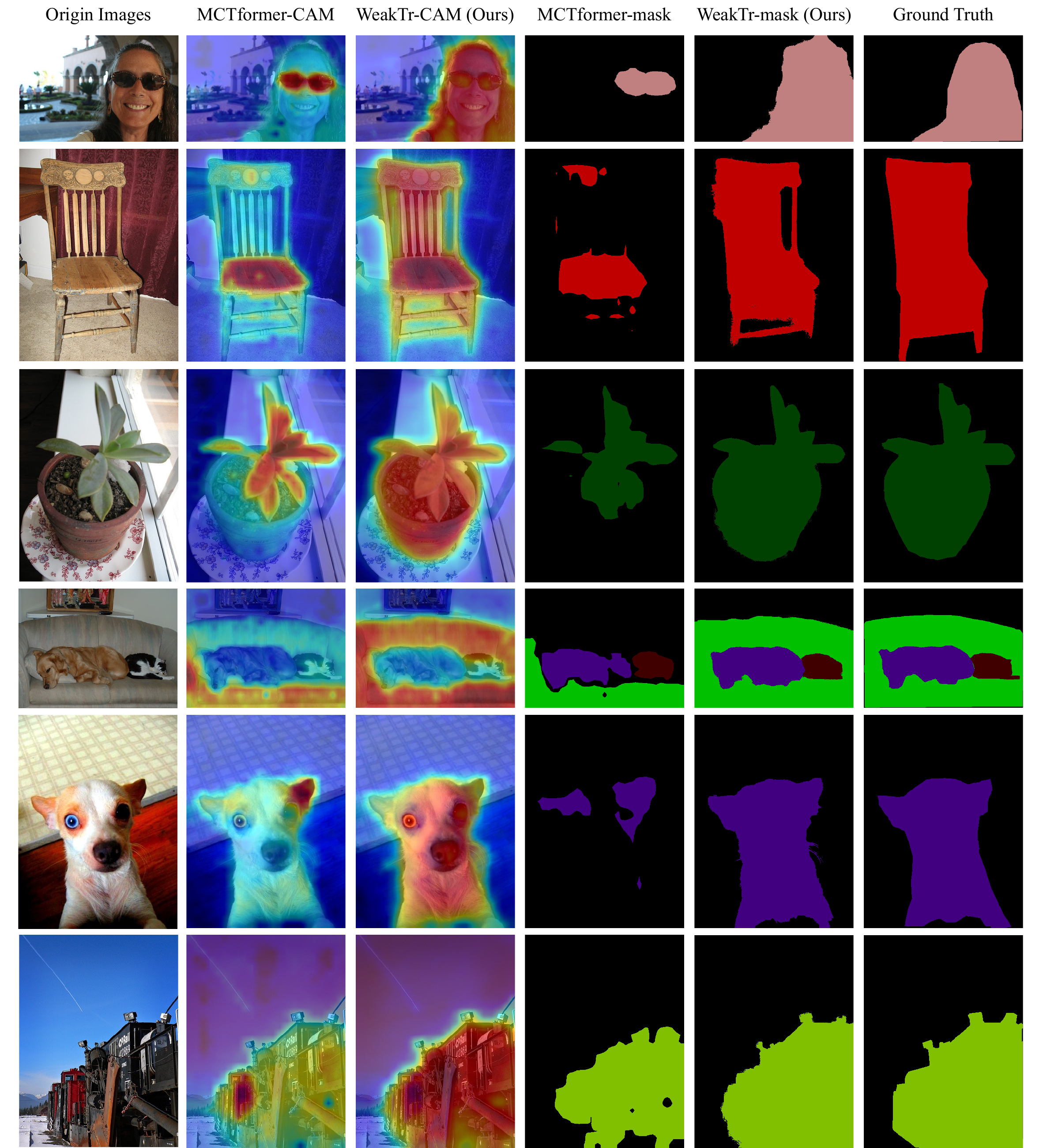}
    \vspace{-.1 in}
    \caption{Comparison of the class activation maps (CAM) on the PASCAL VOC 2012 $train$ set.}
    \label{fig:camcom}
\vspace{-0.2 in}
\end{figure*}

As shown in Fig.~\ref{fig:camcom}, we make a comparison with the MCTformer \cite{xu2022mctformer} for the CAM results. It can be seen that the CAM generated by our WeakTr is more effective than the CAM generated by the MCTformer in terms of generating a high activation response to the entire foreground object. This proves that the weight-based method of WeakTr for the CAM generation can make better use of the plain ViT's self-attention maps for mining the whole object.

\subsubsection{Attention and Activation Results}

\begin{figure*}[htp]
    \renewcommand{\thefigure}{A\arabic{figure}}

    \centering
    \includegraphics[width=1.0\linewidth]{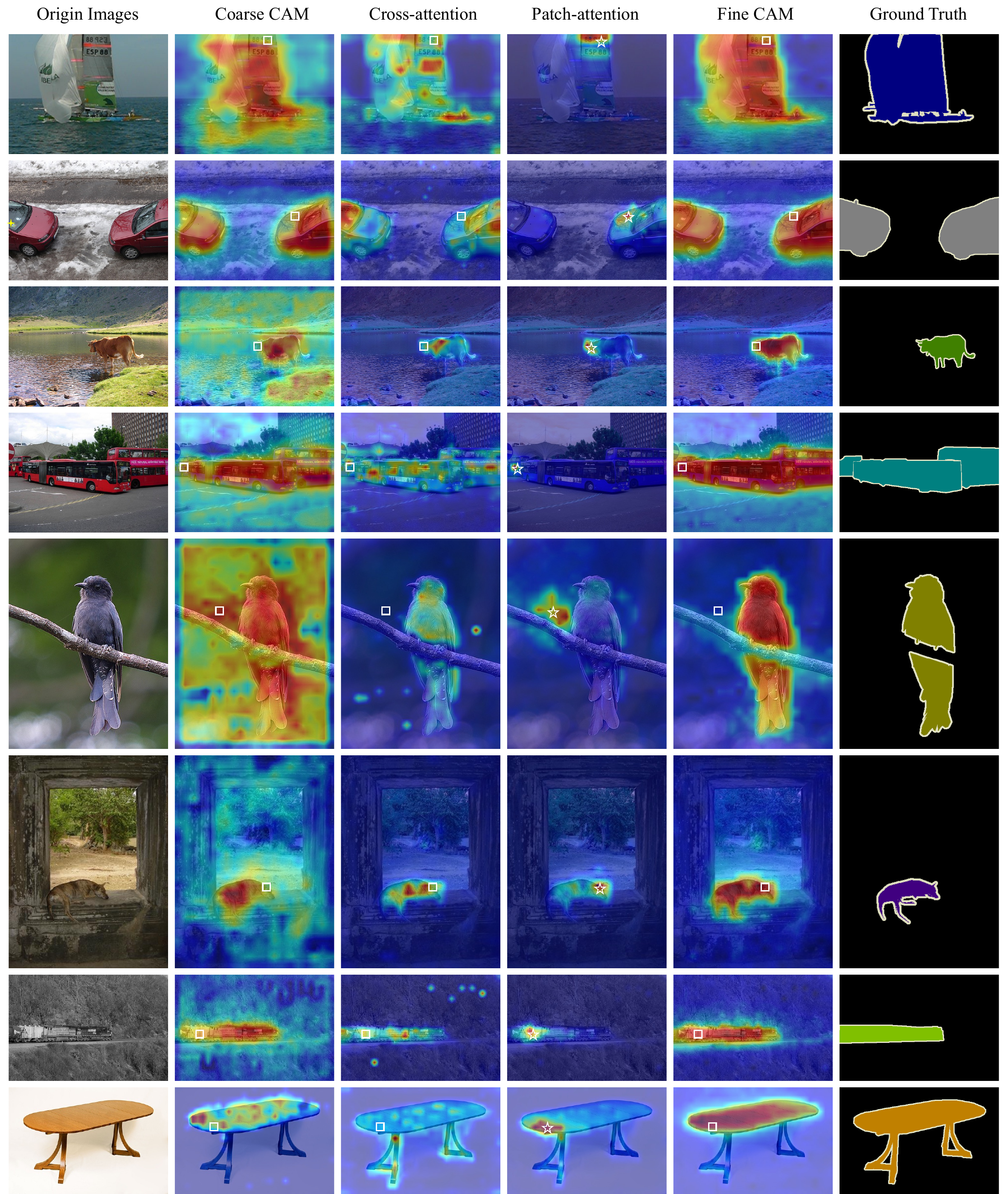}
    \vspace{.15cm}
    \caption{The coarse CAM, the cross-attention, the patch-attention, and the fine CAM results on the PASCAL VOC 2012 $train$ set. We use ``$\star$'' to denote the query points for the patch-attention. We also use ``$\square$'' to indicate the position of query points on the coarse CAM, the cross-attention, and the fine CAM. }
    \label{fig:attncam}
\end{figure*}

As shown in Fig.~\ref{fig:attncam}, we also present the coarse CAM, the cross-attention, the patch-attention, and the fine CAM results on the PASCAL VOC 2012 $train$ set. We can observe that the coarse CAM is usually noisy, while the cross-attention tends to capture only partial object details and sometimes includes noise in the background areas. Patch-attention, on the other hand, typically plays a corrective role for the coarse CAM and cross-attention in local areas. If the activation value of the foreground area is low, the corresponding patch-attention, which contains the attention relationship with the surrounding foreground areas, can be used to increase the activation value. Conversely, if the activation value of the background area is high, the corresponding patch-attention, which contains the attention relationship with the surrounding background areas, can be used to reduce the activation value.

\subsubsection{Semantic Segmentation Results}

\begin{figure*}[htp]
    \renewcommand{\thefigure}{A\arabic{figure}}
    \centering
    \includegraphics[width=0.9\linewidth]{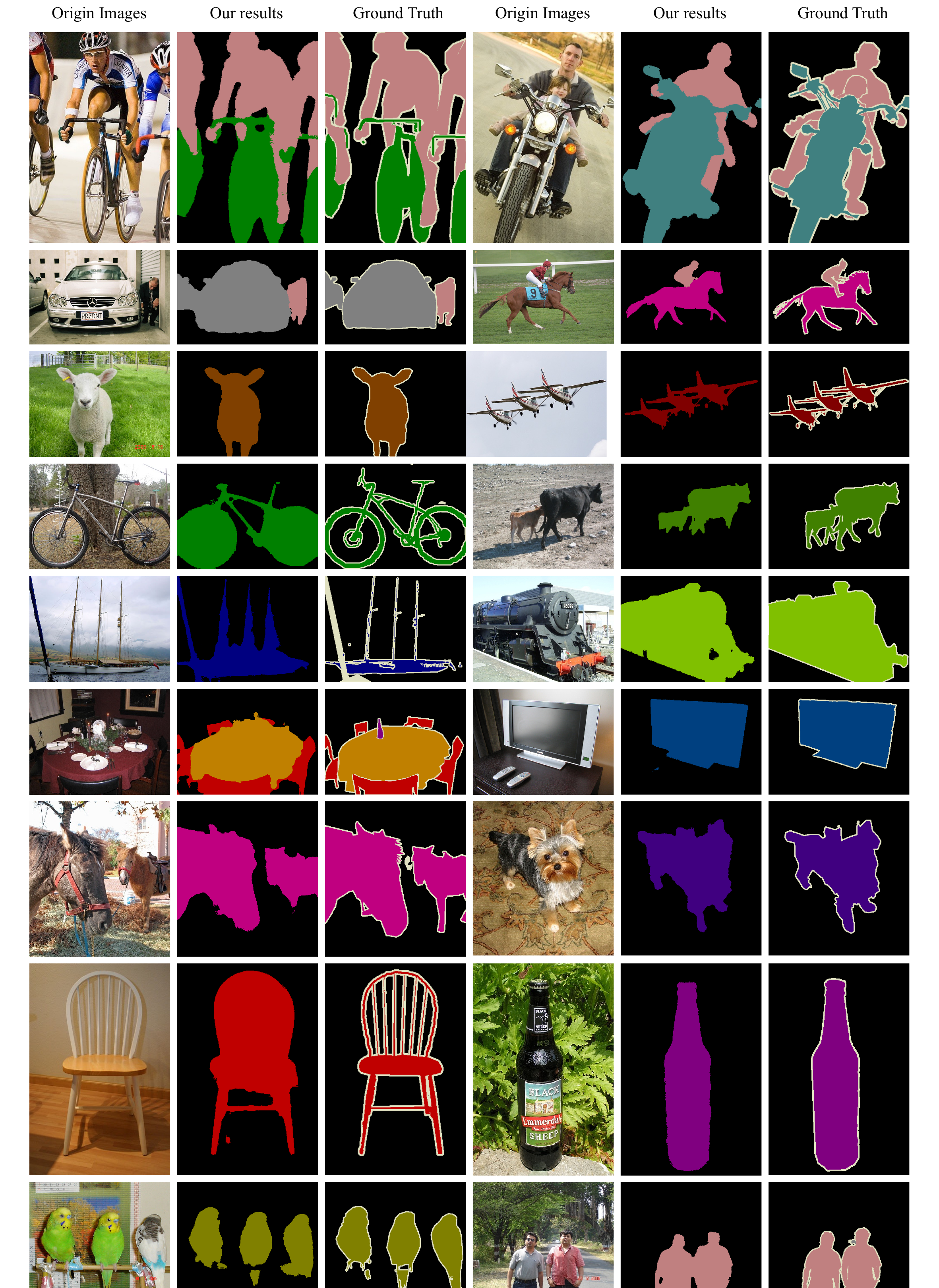}
    \vspace{.2cm}
    \caption{Segmentation visualization results on the PASCAL VOC 2012 $val$ set.}
    \label{fig:visualcom}
\end{figure*}

\begin{figure*}[htp]
    \renewcommand{\thefigure}{A\arabic{figure}}
    \centering
    \includegraphics[width=0.9\linewidth]{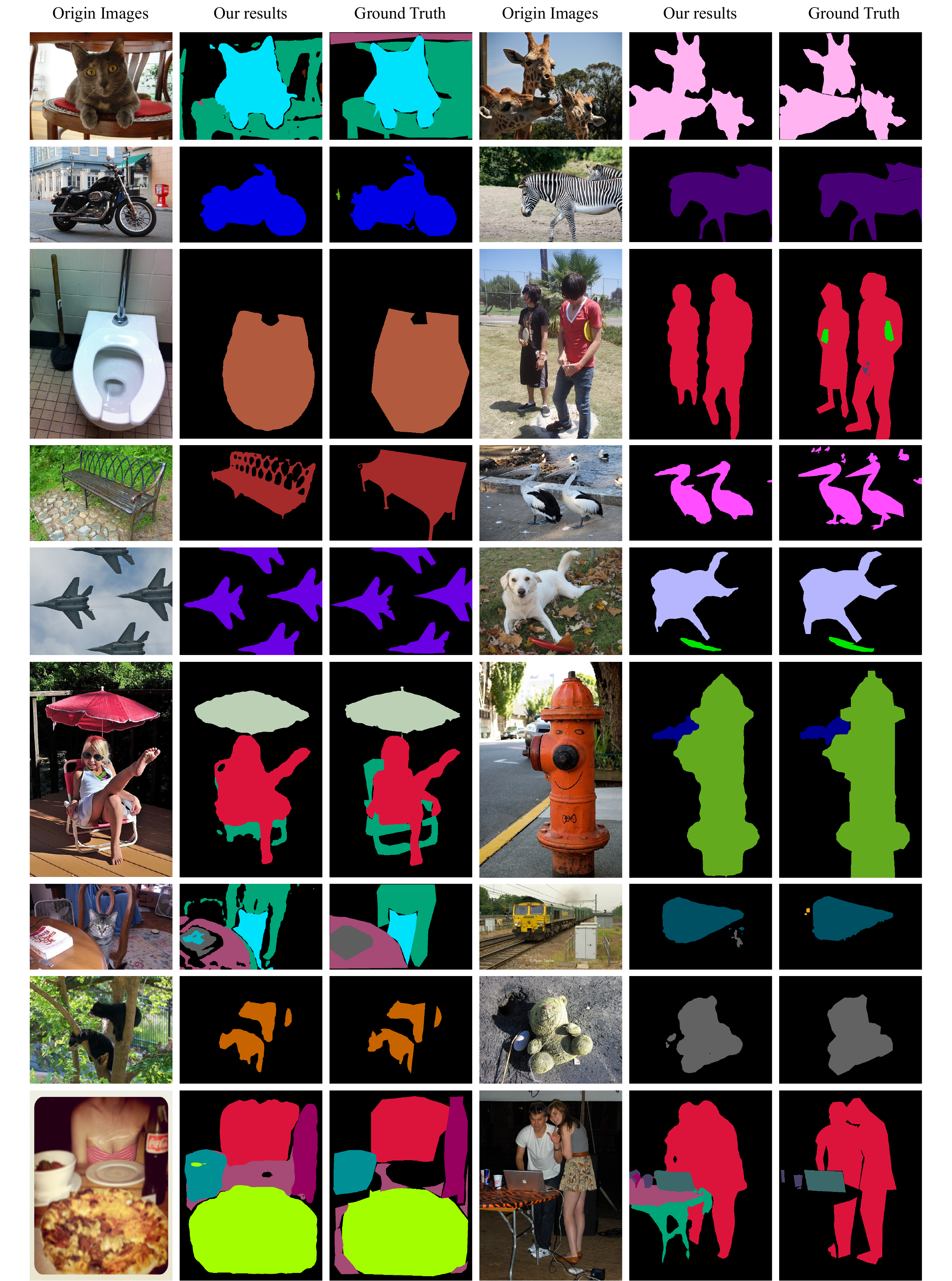}
    \vspace{.2cm}
    \caption{Segmentation visualization results on the COCO 2014 $val$ set.}
    \label{fig:visualcom-coco}
\end{figure*}

We provide the more qualitative segmentation visualization results on the PASCAL VOC 2012 $val$ set in Fig.~\ref{fig:visualcom} and the COCO 2014 $val$ set in Fig.~\ref{fig:visualcom-coco}. We present the original images, our WeakTr segmentation results, and the ground truth (GT). We can observe that for both indoor and outdoor scenes, our WeakTr can provide well-defined segmentation results. Especially for the more complex scenes in the COCO14 dataset, our WeakTr can also give reasonable segmentation results. At the same time, WeakTr also performs well when dealing with obscured objects. The segmentation results demonstrate that WeakTr's online retraining with a gradient clipping decoder can effectively utilize CAM seeds to train the plain ViT-based segmentation network. It also demonstrates that plain ViT-based WSSS has great potential.

\revision{
As shown in Fig.~\ref{fig:visualcom-coco}, we provide representative failure cases for \textbf{extremely small objects} and \textbf{thin parts}. 
For extremely small objects, the predictions may miss the object entirely or merge it into the background category, mainly because the weak image-level supervision and text-guided priors provide limited pixel-level evidence when the object occupies only a few pixels, making the learning easily dominated by contextual cues. 
For thin structures, e.g., the back of chairs, we find the proposed WeakTr shows the ability to predict thin parts while the ground truth ignores the thin part details. 
But the predictions may become broken or overly thickened, since these regions are sensitive to minor localization errors and are difficult to recover from coarse pseudo ground truth.
}

\revision{
We believe these limitations are promising directions for future research. Possible improvements include: 
(i) incorporating higher-resolution and boundary-aware representations, e.g., adding explicit boundary refinement or edge-aware losses, to better preserve thin structures; 
(ii) introducing stronger instance priors, e.g., SAM-style mask proposals or objectness cues, to reduce context dominance and improve small-object recall. 
}

\bibliography{weaktr}

\bibliographystyle{IEEEtran}

\vspace{-15pt}
\begin{IEEEbiography}[{\includegraphics[width=1in,height=1.25in,clip,keepaspectratio]{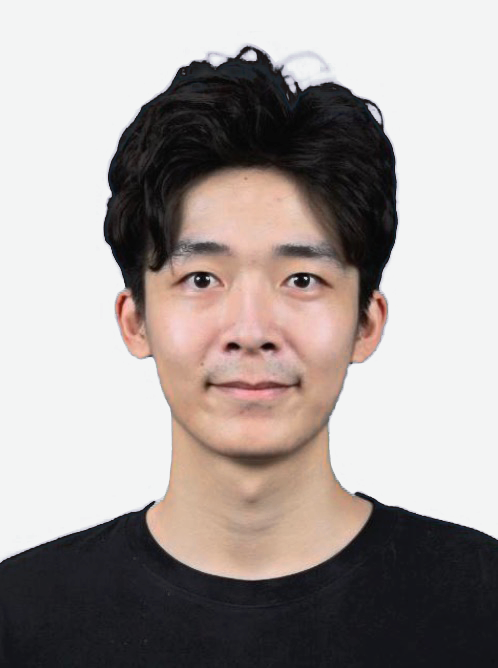}}]{Lianghui Zhu}
    received the B.E. degree from School of Electronic Information and Communications, Huazhong University of Science and Technology, Wuhan, China, in 2021. He is currently a PhD candidate at School of Electronic Information and Communications, Huazhong University of Science and Technology. His research interests include semantic segmentation and weakly-supervised learning.
\end{IEEEbiography}

\vspace{-15pt}
\begin{IEEEbiography}[{\includegraphics[width=1in,height=1.25in,clip,keepaspectratio]{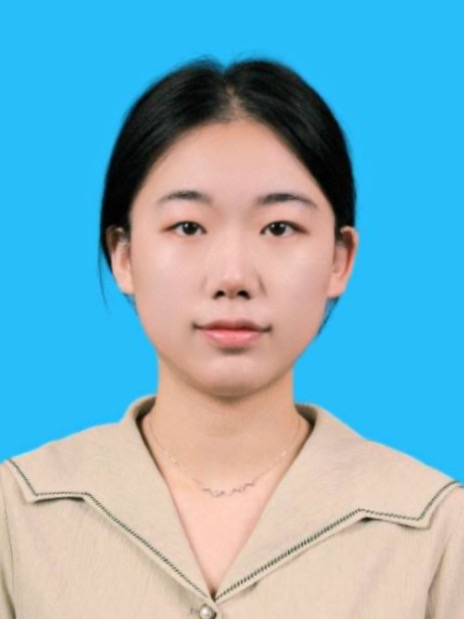}}]{Yingyue Li}
    received the B.E. degree from School of Cyber Science and Technology, Huazhong University of Science and Technology, Wuhan, China, in 2023. She is currently a master candidate at School of Electronic Information and Communications, Huazhong University of Science and Technology. Her research interests include semantic segmentation and weakly-supervised learning.
\end{IEEEbiography}

\vspace{-15pt}
\begin{IEEEbiography}[{\includegraphics[width=1in,height=1.25in,clip,keepaspectratio]{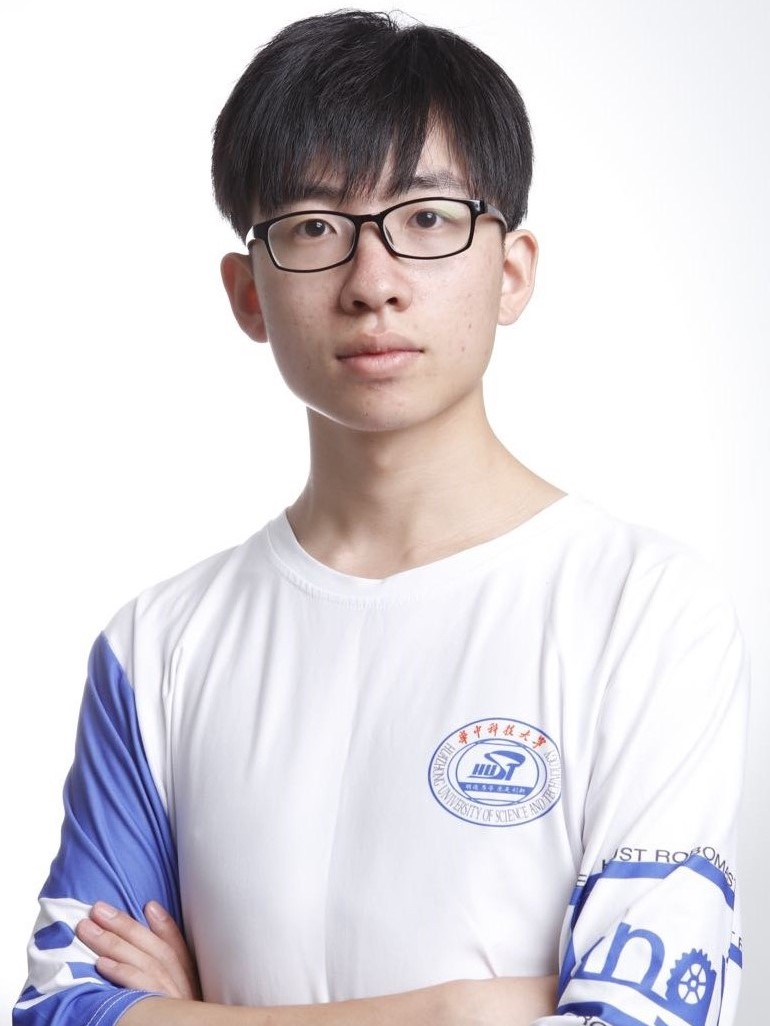}}]{Jiemin Fang}
    received the B.E. and Ph.D. degrees from the School of Electronic Information and Communications, Huazhong University of Science and Technology, in 2018 and 2023 respectively. He is currently a senior researcher at Huawei Inc. His research interests include neural rendering, AutoML, and efficient deep learning. 
\end{IEEEbiography}

\vspace{-15pt}
\begin{IEEEbiography}[{\includegraphics[width=1in,height=1.25in,clip,keepaspectratio]{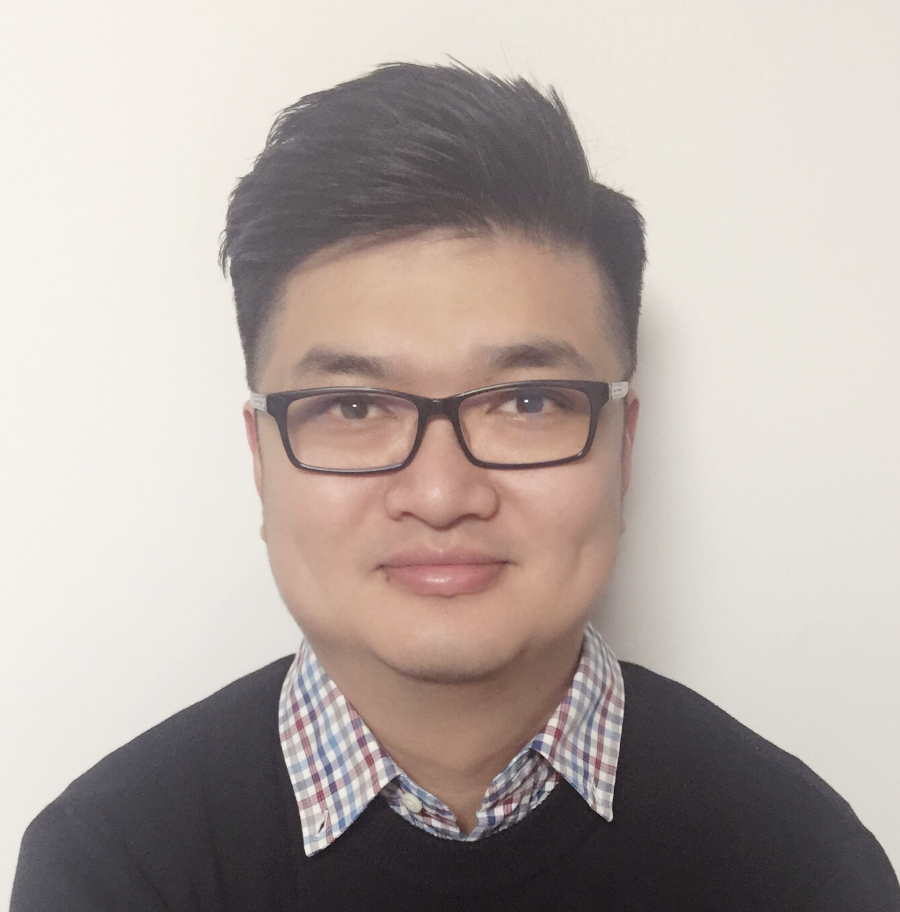}}]{Yan Liu}
    is a researcher and director of Ant Group. Before that, he worked as a senior researcher at Baidu, in charge of the AI security team. His research is on Trusted AI, Networking Security, and Private Computing. He works on broad applications of machine learning and privacy computing technologies in enterprise security.
\end{IEEEbiography}

\vspace{-15pt}
\begin{IEEEbiography}[{\includegraphics[width=1in,height=1.25in,clip,keepaspectratio]{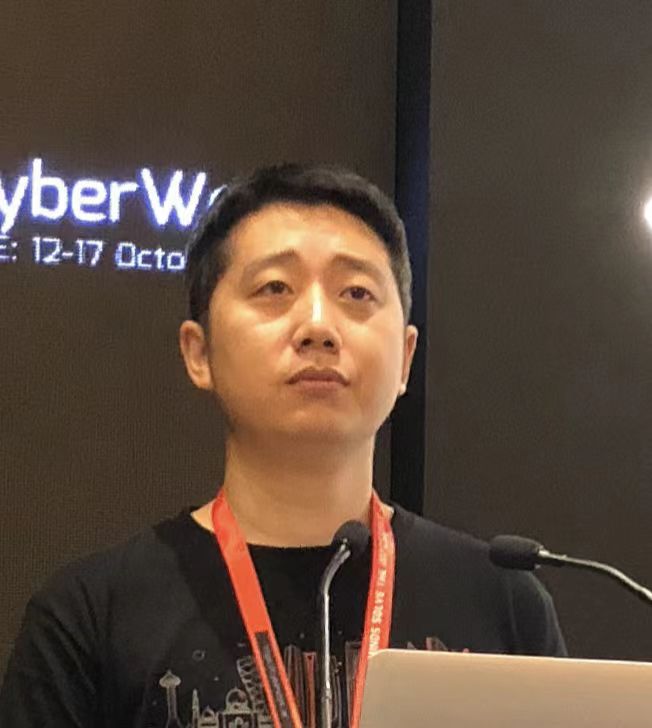}}]{Xin Hao}
    received the B.S. degree in Computer Technology from Harbin Institute of Technology, China, in 2005. He has been engaged in enterprise product development and technical research related to network security, data security, and AI security for many years. Currently working at Ant Group responsible for data security technology research and development.
\end{IEEEbiography}

\vspace{-15pt}
\begin{IEEEbiography}[{\includegraphics[width=1in,height=1.25in,clip,keepaspectratio]{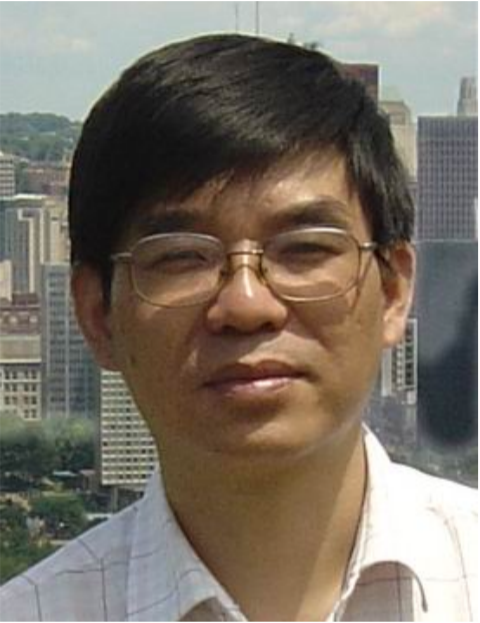}}]{Wenyu Liu (SM’15)}
    received the B.S. degree in Computer Science from Tsinghua University, Beijing, China, in 1986, and the M.S. and Ph.D. degrees, both in Electronics and Information Engineering, from Huazhong University of Science and Technology (HUST), Wuhan, China, in 1991 and 2001, respectively. He is now a professor at the School of Electronic Information and Communications, HUST. His current research areas include computer vision, multimedia, and machine learning.
\end{IEEEbiography}

\vspace{-15pt}
\begin{IEEEbiography}[{\includegraphics[width=1in,height=1.25in,clip,keepaspectratio]{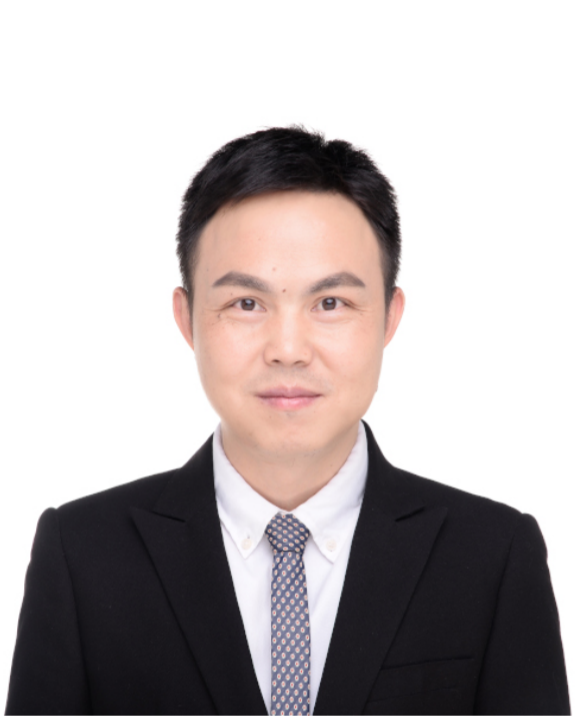}}]{Xinggang Wang}
    received the B.S. and Ph.D. degrees in Electronics and Information Engineering from Huazhong University of Science and Technology (HUST), Wuhan, China, in 2009 and 2014, respectively. He is currently a Professor at the School of Electronic Information and Communications, HUST. He serves as Co-Editor-in-Chief of Image and Vision Computing, associate editor of Pattern Recognition, and area chair of CVPR and ICCV. His research interests include computer vision and deep learning.
\end{IEEEbiography}

\vfill

\end{document}